\documentclass[11pt]{article}
\usepackage{cuted}
\usepackage{listings}
\usepackage{xcolor}
\usepackage[final]{acl}
\usepackage{times}
\usepackage{latexsym}
\usepackage[T1]{fontenc}
\usepackage[utf8]{inputenc}
\usepackage{helvet}

\usepackage{microtype}
\usepackage{inconsolata}
\usepackage{graphicx}
\usepackage{times}
\usepackage{latexsym}
\usepackage[T1]{fontenc}
\usepackage[utf8]{inputenc}
\usepackage{microtype}
\usepackage{graphicx}
\usepackage{array}
\usepackage{ragged2e}
\usepackage{xcolor}
\usepackage{longtable}
\usepackage{array}

\definecolor{LowBlue}{RGB}{33,113,181}
\definecolor{AuthorityBlue}{RGB}{33,113,181}
\definecolor{GenderPurple}{RGB}{106,81,163}
\definecolor{UncertaintyOrange}{RGB}{217,95,2}
\definecolor{TimeGreen}{RGB}{27,120,55}
\definecolor{WellbeingTeal}{RGB}{0,109,119}
\definecolor{FamilyBrown}{RGB}{140,81,10}
\definecolor{KnowledgeRed}{RGB}{178,24,43}
\definecolor{BodyPink}{RGB}{197,27,125}
\definecolor{CommunityGray}{RGB}{64,64,64}
\definecolor{MentalBlue}{RGB}{44,123,182}

\definecolor{nonmedical}{RGB}{33,102,172}
\definecolor{medical}{RGB}{0,128,96}
\definecolor{closed}{RGB}{178,24,43}

\usepackage{xspace,mfirstuc,tabulary}
\usepackage[utf8]{inputenc}
\usepackage[most]{tcolorbox}  

\usepackage{pifont} 
\usepackage{todonotes} 
\usepackage{bbding}
\usepackage{soul} 
\usepackage{xcolor,colortbl} 
\usepackage{appendix}

\usepackage{enumitem} 
\usepackage{booktabs} 
\usepackage{algorithm} 
\usepackage{multirow} 
\usepackage{array} 
\usepackage{arydshln} 
\usepackage{amsmath, amssymb} 
\usepackage{algpseudocode} 
\usepackage{hyperref}
\usepackage[table]{xcolor}
\usepackage[normalem]{ulem}
\usepackage{multirow}
\usepackage[table]{xcolor}

\usepackage{tabularx}
\definecolor{ForestGreen}{rgb}{0.13, 0.55, 0.13}
\definecolor{darkred}{RGB}{139, 0, 0}
\definecolor{humanrow}{RGB}{230,245,235}

\newcommand{\framework}{\texttt{MH-INDIC}}
\newtcolorbox{promptbox}[2][]{%
  breakable,
  colback=black!2,
  colframe=black,
  title={#2},
  fonttitle=\bfseries,
  sharp corners,
  boxrule=0.6pt,
  left=6pt,right=6pt,top=6pt,bottom=6pt,
  #1
}
\usepackage{listings}
\usepackage{xcolor}

\title{Measuring Cultural Alignment Beyond the Average: A Framework for Evaluating Maternal-Health LLM Interactions in Indian Contexts}

\author{Umaira Izhar, Gunjan Arora, \textbf{Pushpendra Singh}\\
  Indraprastha Institute of Information Technology Delhi;\\
  \texttt{\{umairai,gunjana,psingh\}@iiitd.ac.in}\\ 
}

\begin{document}
\maketitle{}
\begin{abstract}
Existing evaluation methods for healthcare LLMs primarily assess factual correctness, safety, and fluency, while providing limited insight into whether generated interactions reflect culturally situated healthcare reasoning. This limitation is particularly important in maternal health, where care decisions are shaped by social and relational norms. We introduce MH-INDIC, a culturally grounded evaluation framework for maternal-health interactions in urban and semi-urban North Indian contexts that operationalises cultural behaviour through ten dimensions of maternal-health reasoning. Using a 26-item survey administered to 102 pregnant and postpartum women from urban and semi-urban North India, we evaluate ten LLMs. We distinguish population-level cultural alignment from profile-level behavioural variation. Although several models approximate the human population-level distribution, all evaluated systems exhibit substantially lower variation across demographic and household profiles than the human cohort, revealing a gap between aggregate alignment and profile-conditioned sensitivity. As a downstream application of MH-INDIC, we use the strongest-aligned proprietary and open-source models to generate culturally conditioned maternal-health dialogues under zero-shot, self-conditioned, and human-grounded prompting. Human-grounded conditioning produces stronger profile alignment and dialogue-quality ratings, suggesting that measured cultural profiles can improve the cultural grounding of generated interactions.
\end{abstract}
\maketitle

\begin{figure}[t]
    \centering
    \includegraphics[width=0.7\columnwidth]{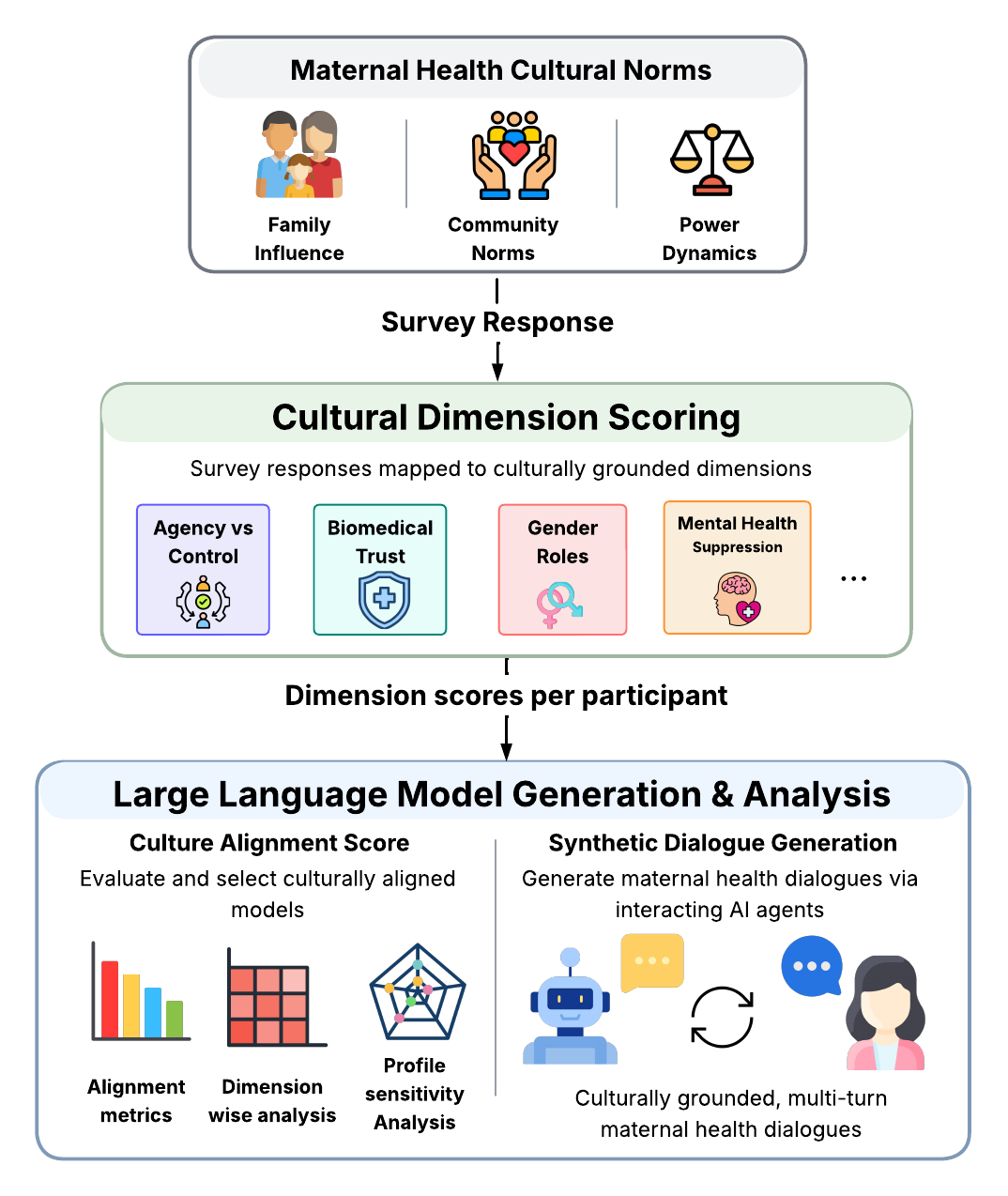}
    \caption{Conceptual overview of MH-INDIC. Survey responses are mapped to culturally grounded dimensions, used to evaluate cultural alignment in LLMs, and applied to generate culturally situated maternal health dialogues.}
    \label{fig:example}
\vspace{-2.5mm}
\end{figure}

\section{Introduction}

Recent advances in Large Language Models (LLMs) have enabled their use in privacy-sensitive and low-resource settings~\cite{suresh2025diasynthsyntheticdialoguegeneration,kang2024syntheticdatagenerationllm}. Prior work has explored synthetic conversational data generation across diverse application domains, including healthcare and education~\cite{lee2024cactuspsychologicalcounselingconversations,chitale2026updeshsynthesizinggroundedinstruction, Das2024SyntheticPD}. These systems have demonstrated strong performance in conversational fluency, contextual relevance, and medically grounded response generation, helping mitigate data scarcity in privacy-sensitive environments.

A growing body of this work targets maternal health, where LLM-based assistants are being explored for antenatal information delivery, postpartum support, and training community health workers in low and middle-income contexts~\cite{zhang2026cama, Ramjee_2025,10.1145/3706598.3713362}. However, maternal health decisions are rarely negotiated solely between the patient and the clinician. Instead, they emerge within extended networks involving spouses, elders, community health workers, and informal knowledge sources~\cite{shibeshi2023understanding}. Within these settings, norms surrounding deference to authority, attitudes toward medical uncertainty, emotional disclosure, and moral expectations of self-sacrifice substantially influence women's access to care and their lived experiences during pregnancy and the postpartum period.

These social and normative dimensions are difficult to capture with existing evaluation methods. Existing approaches for evaluating LLM-generated healthcare interactions primarily emphasise medical accuracy, safety, linguistic fluency, and broad cultural appropriateness~\cite{kazemi2024culturalfidelitylargelanguagemodels,rystrom2025multilingual}. Meanwhile, Hofstede-inspired frameworks~\cite{hofstede2011dimensionalizing} and recent LLM-oriented cultural benchmarks~\cite{zhao2024worldvaluesbenchlargescalebenchmarkdataset,chiu2024culturalbench,li2024cultureparkboostingcrossculturalunderstanding} remain too coarse to capture the socially situated dynamics underlying maternal-health behaviour in low-resource contexts. As a result, they provide limited insight into whether LLMs reproduce the reasoning through which healthcare decisions are negotiated and acted upon.

To address this limitation, we propose \framework {}, a culturally grounded evaluation framework for maternal-health LLM interactions in urban and semi-urban North Indian contexts. To the best of our knowledge, this is the first work to systematically evaluate cultural alignment in such interactions using a multidimensional framework. \framework {} builds on Hofstede's dimensional theory of culture~\cite{hofstede2011dimensionalizing}, while extending it with maternal-health-specific relational and social dimensions better suited to low-resource care contexts. The resulting ten-dimensional framework captures medical and familial authority, gendered caregiving expectations, comfort with medical uncertainty, temporal orientation, maternal self-care and emotional needs, community conformity, maternal agency, biomedical versus traditional care beliefs, bodily disclosure, and mental-health suppression. The full set of dimensions and their maternal-health interpretations are described in Section~\ref{sec:mhindic}.

We operationalise \framework{} as a 26-item survey instrument
developed in consultation with mental and maternal-health experts
(Appendix~\ref{app:possitionality}) and grounded in prior
literature~\cite{10.1145/3706598.3713362,kaur2025did,hamal2020social,zhang2026cama},
and collect responses from 102 pregnant and postpartum women in
urban and semi-urban North India to establish the empirical human
reference. We then evaluate ten LLMs across proprietary, open-source, and medical-domain families using profile-conditioned prompting, in which models simulate responses conditioned on demographic and household profiles. Each model is tested across 30 maternal profiles sampled from the cohort. An identical rule-based scoring framework is applied to human and model responses, enabling systematic alignment analysis across all ten dimensions.

We further examine the implications of cultural alignment for downstream generation by producing synthetic maternal-health dialogues under self-conditioned and human-grounded prompting regimes. Using the strongest aligned open-source and proprietary models identified in the survey evaluation, we generate dialogues for three cohort-derived maternal profiles and two synthetic profiles. Through blinded expert evaluation, we assess the cultural plausibility, affective realism, and demographic consistency of the resulting conversations, linking representation-level alignment to interactional quality. Together, these components provide a unified evaluation pipeline for cultural alignment in maternal-health LLMs across survey-level evaluation and downstream dialogue generation. We summarise our contributions as follows:

\begin{itemize}[leftmargin=*, noitemsep]

\item We introduce \framework {}, a multidimensional evaluation framework and survey-grounded benchmark for maternal-health LLM interactions in in urban and semi-urban North Indian contexts.

\item We propose a marginal-conditional decomposition of alignment that separates population-level agreement from sensitivity to profile-level behavioural variation.

\item Through systematic evaluation across proprietary, open-source, and medical-domain LLMs, we show that strong aggregate alignment does not necessarily imply human-like behavioural sensitivity, and that stronger alignment improves the realism of downstream synthetic maternal-health dialogues.

Code for scoring, LLM survey responses, and dialogue generation is available at: \href{https://anonymous.4open.science/r/MH-INDIC-0298/}{GitHub}

\end{itemize}

\section{Related Work}

Maternal health research in public health and social science has established that cultural norms, family structures, and authority relations strongly shape care-seeking behaviour and maternal outcomes~\cite{yadav2022should,10.1145/3706598.3713362}. As AI systems are increasingly deployed in healthcare-facing settings, cultural alignment has become critical, particularly in socially embedded and high-stakes domains. Our work builds on three research strands: AI applications in maternal and reproductive health, cultural modelling and evaluation in NLP, and culturally grounded dialogue generation.

Prior work has applied AI to risk prediction, clinical decision support, and informational chatbots for maternal health~\cite{Ramjee_2025,islam2022machine}. While effective in improving access and workflow efficiency, most approaches primarily model biomedical or individual-level factors, with limited representation of broader socio-cultural context. Key influences such as spousal authority, elder roles, and community norms are often abstracted or ignored, limiting these systems' ability to capture the relational and constraint-driven nature of maternal decision-making~\cite{kaur2025did}. A small but growing line of work has begun to engage cultural sensitivity in maternal and reproductive-health chatbots~\cite{10.1145/3706598.3713362}, but does not yet provide a generalisable evaluation framework for assessing whether such systems reproduce culturally situated reasoning at scale.

Recent NLP research has examined cultural bias, cross-cultural variation, and value alignment in language models, proposing benchmarks for cultural awareness and moral reasoning~\cite{tao2024cultural,li2024cultureparkboostingcrossculturalunderstanding}. However, these frameworks typically rely on abstract value categories and crowd-sourced judgments, lacking grounding in domain-specific cultural structures~\cite{zhao2024worldvaluesbenchlargescalebenchmarkdataset}. In contrast, maternal-health decisions are deeply relational and context-dependent, shaped by hierarchical authority and situational constraints, motivating evaluation frameworks that capture how values operate under real-world decision pressures. Existing benchmarks also typically report alignment at the population-marginal level, how close a model's mean responses sit to the human mean, but rarely examine whether models reproduce culturally meaningful variation across individuals, households, and demographic profiles~\cite{mukherjee2024culturalconditioningplaceboeffectiveness}.

Dialogue-generation research has explored personalisation, controllability, and value alignment, including empathetic and supportive health dialogue~\cite{zhang2026cama,lee2024cactuspsychologicalcounselingconversations}. However, cultural grounding is often approximated through surface-level conditioning, without explicit evaluation of whether generated responses reflect underlying sociocultural constraints. Moreover, few studies assess whether models demonstrate consistent cultural reasoning across both evaluation and generation tasks~\cite{chiu2024culturalbench}, a gap our framework directly addresses.

In summary, prior work has studied maternal health, cultural modeling, and dialogue generation largely in isolation. We bridge this gap through a culturally grounded evaluation framework that operationalises maternal-health behaviour via empirically defined cultural dimensions, enables direct comparison between human and model behaviour, and connects survey-level cultural alignment with downstream dialogue generation. This enables principled evaluation of whether language models genuinely represent and consistently apply culturally situated reasoning, rather than producing only superficially appropriate responses.

\section{MH-INDIC}
\label{sec:mhindic}

Our approach builds on Geert Hofstede's theory of culture, which models culture through interpretable social and behavioural dimensions~\cite{hofstede2011dimensionalizing}. We use Hofstede's dimensional methodology as an interpretable scaffold rather than adopting its national-level cultural scores directly. Its dimensions provide established axes for representing variation in authority, uncertainty management, temporal orientation, gendered expectations, self-regulation, and individual–collective decision-making. However, Hofstede's framework primarily operates at national and population levels, whereas maternal-health interactions involve socially situated forms of decision-making shaped by relationships among mothers, clinicians, spouses, elders, and communities. We therefore reinterpret all six Hofstede dimensions at the maternal-health interaction level and introduce four additional dimensions grounded in maternal-care and public-health literature, yielding a ten-dimensional framework.
\\
\\
\textbf{Dimensions Overview:} Table~\ref{tab:maternal_culture_dimensions} summarises the culturally grounded dimensions used in our framework and their maternal-care interpretations. We reinterpret \textit{Power Distance Index (PDI)} as navigation of medical and institutional authority, \textit{Uncertainty Avoidance Index (UAI)} as operationalising comfort with medical ambiguity and unfamiliar care procedures, \textit{Long-Term vs Short-Term Orientation (LTO)} as prioritisation of future maternal and child health, \textit{Indulgence versus Restraint (IVR)} as attitudes toward maternal self-care, comfort, rest, and the expression of personal distress. \textit{Masculinity vs Femininity
} renamed as \textit{Gender Norms (GN)}, captures gendered expectations surrounding motherhood and caregiving and \textit{Individualism versus Collectivism}  renamed as \textit{Community Norms versus Individual Divergence (CNID)}, capturing tensions between social conformity and individual maternal decision-making. 

Additional domain-specific dimensions include \textit{Maternal Agency versus Familial Control (MAFC)}, which captures the degree of maternal autonomy within family-centred decision-making; \textit{Biomedical Trust versus Traditional Belief (BTB)}, reflecting preferences between biomedical and traditional care systems; \textit{Silence versus Openness around the Maternal Body (SVO)}, capturing norms around bodily disclosure and symptom expression; and \textit{Mental Health Suppression (MHS)}, reflecting recognition and expression of maternal emotional distress. Together, these dimensions form the conceptual backbone of \framework, enabling systematic evaluation of culturally grounded maternal-health behaviour.

\begin{figure*}[t]
    \centering
    \includegraphics[width=0.9\textwidth]{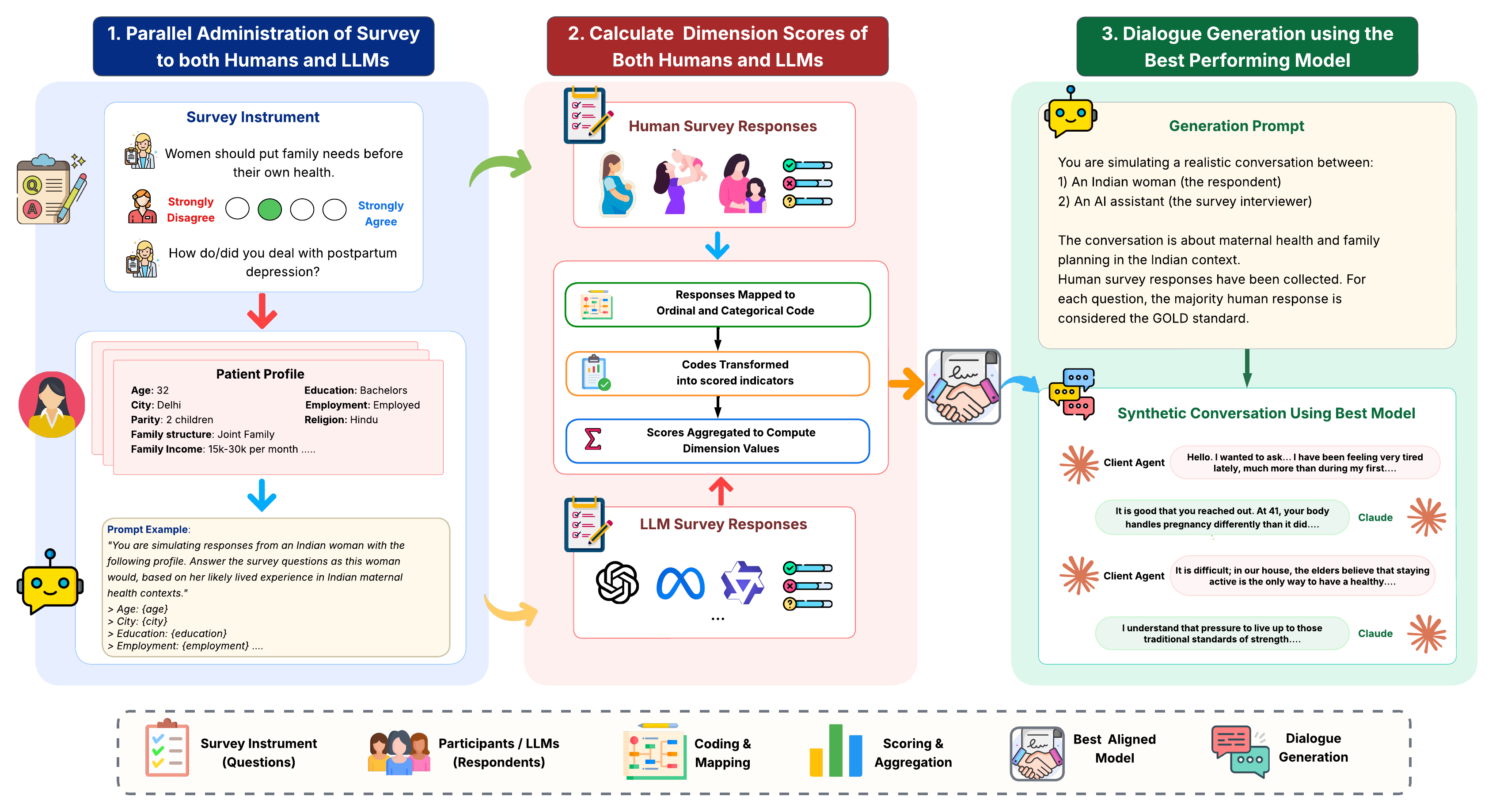}
    
    \caption{Overview of the \framework{} pipeline. Human participants and LLMs complete matched maternal health surveys, whose responses are converted into normalised cultural indicators and aggregated into dimension-level representations. Human responses serve as the reference for evaluating LLM cultural alignment and selecting the best-aligned model, which is then used for culturally conditioned maternal health dialogue generation.
}
    \label{fig:main_framework}
    \vspace{-2mm}
\end{figure*}

\section{Methodology}

We propose a structured evaluation pipeline to assess the cultural alignment of LLMs in maternal health contexts and to generate culturally conditioned maternal health dialogues. The pipeline consists of four stages: (i) survey design, (ii) parallel administration to humans and LLMs, (iii) behavioural indicator construction and scoring, (iv) culturally grounded dialogue generation. Figure \ref{fig:main_framework} illustrates the entire pipeline.

\subsection{Survey Design and Administration}

We design a structured survey to operationalise culturally salient
aspects of maternal-health decision-making across pregnancy,
childbirth, and the postpartum period. The instrument comprises
Likert-scale, multiple-choice, and open-ended items mapped to the
ten \framework{} dimensions, with each construct represented through
complementary direct and social-contextual indicators. The instrument was developed through review of maternal-health and public-health literature and refined through consultation with two domain experts: a maternal-health expert (MBBS/MPH) and a mental-health expert (MSc Psychology and practicing counsellor). The experts reviewed the construct definitions, item wording, and item-dimension mappings. The survey was administered in English.

We assess the structural validity of \framework{} via psychometric
analysis on human responses, including Cronbach's $\alpha$, item-total
correlations, EFA loadings, and inter-dimension correlations. Because several dimensions are behaviourally heterogeneous and formative rather than reflective, these analyses are used to assess structural interpretability rather than to require high internal consistency across all dimensions. For example, MHS combines complementary indicators of experienced distress, emotional disclosure, and community recognition, which need not be strongly correlated even when they jointly characterise mental-health suppression.
(Appendix~\ref{app:psychometric}).
The survey is administered to 102 pregnant and postpartum women in
urban and semi-urban North India, whose responses form the empirical
human reference (Appendix~\ref{app:demographics},
\ref{app:survey}). The same instrument is then administered to ten
LLMs under matched persona conditioning, enabling directly comparable
evaluation within the same multidimensional space.

\subsection{Behavioural Indicator Construction and Scoring}

All survey responses are converted into dimension-level representations using a predefined rule-based coding framework. Binary responses are encoded as $\{0,1\}$, while ordinal responses are normalised to the range $[0,1]$, with reverse coding applied where necessary to maintain consistent interpretation.

Most dimensions are derived from multiple related survey responses rather than individual items alone. Scoring and aggregation procedures additionally incorporate asymmetric rules to preserve relational dynamics that may otherwise be obscured by simple averaging. A summary of dimension construction and aggregation procedures is provided in Table~\ref{app:survey_dimensions}.

MAFC and MHS are additionally decomposed into interpretable subcomponents. MAFC is separated into spouse-mediated and elder-mediated authority structures, while MHS is decomposed into personal emotional disclosure and community-level recognition of maternal distress. These decompositions are used for secondary analysis while preserving a stable set of primary dimensions for alignment evaluation.

All item-level scores are then aggregated into fixed-length cultural representations across the ten \framework{} dimensions. Identical scoring procedures are applied to both human and model responses, enabling directly comparable alignment analysis. Full coding rules, aggregation procedures, and scoring examples are provided in Appendix~\ref{app:scoring}.

\subsection{Aggregate and Profile-Conditioned Alignment}

We evaluate cultural alignment at both aggregate and profile-conditioned levels. Aggregate alignment measures how closely model responses reproduce population-level cultural tendencies, while profile-conditioned alignment evaluates whether models preserve behavioural variation across demographic and household profiles.

Let $\mu_H \in \mathbb{R}^{|D|}$ denote the mean human cultural representation over primary dimensions $D$, and $\mu_m$ the corresponding representation for model $m$. Aggregate alignment is computed as:
\[
A_m = \|\mu_H - \mu_m\|_2
\]
where lower values indicate stronger alignment with the human reference distribution.

Profile-conditioned sensitivity is measured as:
\[
C_m = \frac{1}{|D|}\sum_{d \in D}\sigma_d^{(m)}
\]
where $\sigma_d^{(m)}$ denotes the standard deviation of model responses for dimension $d$ across profile-conditioned evaluations. Human conditional sensitivity is estimated analogously using subgroup-level demographic and household variation. Analysis is performed on primary dimensions only, with subcomponents decomposed and evaluated separately to avoid overweighting related constructs.




\begin{figure*}[t]
     \centering
     \includegraphics[width=0.8\textwidth]{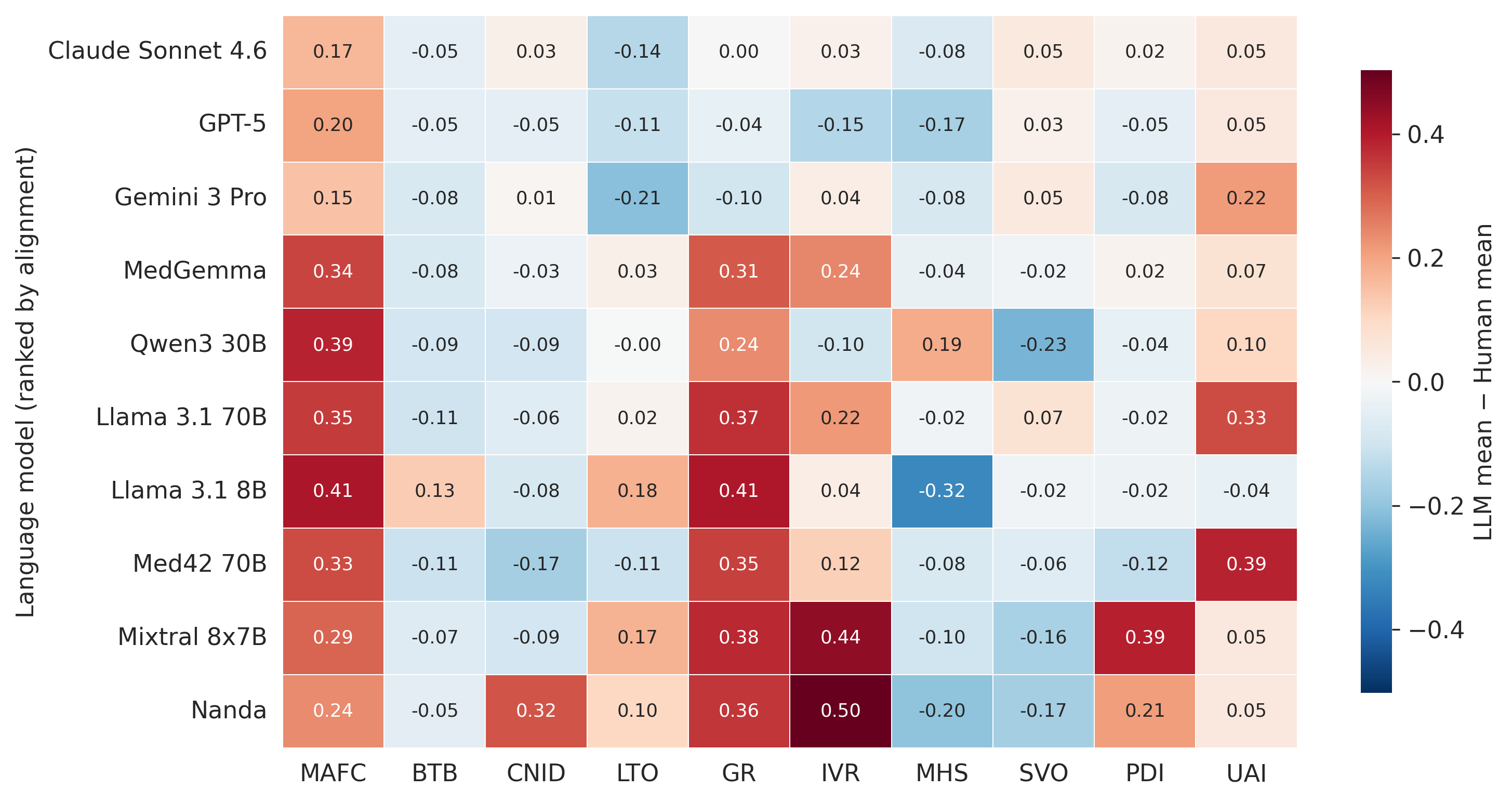}
    
     \caption{Absolute deviation between model and human mean scores across the ten \framework{} dimensions at $T{=}0.3$. Higher values indicate greater cultural misalignment. Models are ordered by overall alignment performance, from the best-aligned (Claude Sonnet 4.6) to the least-aligned (Nanda).
 }
     \label{fig:dimension_heatmap}
     \vspace{-2mm}
 \end{figure*}
 
\subsection{Culturally Grounded Dialogue Generation}

As a downstream application of \framework{}, we use the strongest aligned open-source and proprietary models to generate synthetic maternal-health dialogues under three prompting regimes: zero-shot, self-conditioned, and human-grounded. Conversations are simulated as multi-turn interactions between an AI maternal-health companion and pregnant or postpartum client personas.

Each dialogue is anchored around one primary dimension(PDI), representing the behavioural signal under evaluation, and two secondary dimensions (MAFC and GN), which provide contextual social and relational cues without diluting the primary signal. These secondary dimensions are held fixed across all profiles and conditioning regimes, ensuring that differences in Profile Alignment Score reflect the conditioning strategy rather than dimension selection. In the zero-shot regime, models receive only the maternal profile and target dimensions. Self-conditioned prompting uses behavioural markers derived from the model's own survey responses, while human-grounded prompting conditions generation on human respondent scores for the corresponding profile. Conditioning is implemented through dimension-specific prompts encoding behavioural tendencies, communication styles, family dynamics, and culturally situated decision-making patterns (Appendix~\ref{app:prompts}).

For each profile and prompting regime, we generate paired 20-turn conversations while holding all other prompting conditions constant. The resulting dialogues are evaluated through blinded expert assessment to analyse how different forms of cultural conditioning influence interaction quality.

\section{Experiments}

\subsection{Human-LLM cultural alignment:}
We evaluate ten LLMs spanning proprietary models (Claude Sonnet 4.6, GPT-5, Gemini 3 Pro), open-source models (Llama 3.1 70B, Llama 3.1 8B, Qwen3 30B, Mixtral 8$\times$7B), medical-domain models (MedGemma, Med42 70B), and the Hindi-adapted Nanda 10B model ~\cite{claude2025sonnet, dubey2024llama3,nanda2025hindi,openai2025gpt5,google2025gemini3}. Open-weight models are served locally through vLLM on NVIDIA A100 80GB GPUs, while proprietary systems are accessed through their respective APIs.

Each model is conditioned on maternal profiles drawn from the cohort and evaluated across 30 profiles selected using farthest-first traversal over seven demographic attributes (age, education, employment, family structure, parity, income, and religion). Responses are generated greedily at temperature $T=0.0$ and using majority voting over five samples at $T=0.3$ and $T=0.7$; unless otherwise stated, we report results for $T=0.3$.

\subsection{Dialogue Generation.}
Using Claude Sonnet 4.6 and Qwen3 30B, we generate synthetic maternal-health dialogues. Evaluation is performed on five profiles: three real cohort profiles spanning autonomous (P24), mid-range (P25), and conflicting configurations (P18), together with two synthetic extreme profiles representing fully autonomous (PSYN-LOW) and fully constrained orientations (PSYN-HIGH) (c.f. Appendix~\ref{app:profiles}. This setup evaluates whether models preserve coherent yet diverse cultural postures under varying conditioning strategies.

\subsection{Evaluation Metrics}

For cultural alignment, we compare model-derived \framework{} profiles with the human cohort using Euclidean distance, mean absolute error (MAE), cosine similarity, and Spearman's rank correlation. Per-dimension significance testing is performed using Mann-Whitney U with Benjamini-Hochberg FDR correction ($q=0.05$). We estimate 95\% confidence
intervals for the Euclidean alignment distance by bootstrap resampling
the 102 human profiles with replacement ($B=1000$).
\begin{table}[t]
\centering
\small
\setlength{\tabcolsep}{5pt}
\renewcommand{\arraystretch}{1.08}

\resizebox{\columnwidth}{!}{
\begin{tabular}{lcccc}
\toprule

\textbf{Models} 
& \textbf{Eucl.}$\downarrow$ 
& \textbf{MAE}$\downarrow$ 
& \textbf{Cos.}$\uparrow$ 
& $\boldsymbol{\rho}^{\uparrow}$ \\

\midrule

\multicolumn{5}{l}{\textit{\textbf{Open-Source Models}}} \\
\midrule

Llama 3.1 70B       
& .659 
& .157 
& .916 
& .309 \\

Qwen3 30B           
& .580 
& .148 
& .873 
& .200 \\

Mixtral 8$\times$7B 
& .814 
& .216 
& .876 
& .152 \\

Llama 3.1 8B        
& .704 
& .165 
& .841 
& .139 \\

\midrule

\multicolumn{5}{l}{\textit{\textbf{Medical-Domain Models}}} \\
\midrule

MedGemma  
& .535 
& .118 
& .929 
& .358 \\

Med42 70B 
& .686 
& .183 
& .849 
& .333 \\

\midrule

\multicolumn{5}{l}{\textit{\textbf{Hindi-Adapted Models}}} \\
\midrule

Nanda 10B 
& .817 
& .221 
& .900 
& .406 \\

\midrule

\multicolumn{5}{l}{\textit{\textbf{Proprietary Models}}} \\
\midrule

GPT-5        
& .343 
& .090 
& .932 
& .406 \\

Gemini 3 Pro 
& .380 
& .101 
& .924 
& .358 \\

\rowcolor{green!20}
Claude Sonnet 4.6 
& \textbf{.253} 
& \textbf{.061} 
& \textbf{.968} 
& \textbf{.527} \\
\bottomrule
\end{tabular}
}
\caption{
Human-LLM cultural alignment across the 10 \framework{} dimensions at $T{=}0.3$. Lower is better for Euclidean distance and MAE; higher for cosine similarity and Spearman $\rho$. Best values are in \textbf{bold}.}

\label{tab:aggregate_alignment}

\end{table}
For dialogue evaluation, each generated dialogue is rated on the ten \framework {} cultural dimensions together with three general, dialogue-quality metrics: \textit{Profile Consistency} (whether the dialogue remains coherent with the conditioning profile across all turns), \textit{Naturalness \& Fluency} (whether the dialogue reads as natural human conversation), and \textit{Overall Dialogue Quality} (a holistic judgement of realism, coherence, emotional plausibility, and contextual grounding). Ratings use a 1-5 Likert scale, with orientation dimensions ranging from autonomous to constrained and quality dimensions ranging from poor to excellent. Orientation ratings additionally define a Profile Alignment Score (PAS):
\[
\mathrm{PAS}(c) \;=\; 1 \;-\; \frac{1}{4\,|\mathcal{D}|}
                       \sum_{d \in \mathcal{D}}
                       \bigl|\,y^{*}_{d} - \hat{y}_{d}(c)\,\bigr|,
\]
\\
where \(c\) is the conversation, \(D\) the set of dimensions, \(\hat{y}_d(c)\) is the rated Likert \(c\) on dimension \(d\), and \(y_d^*\) is the target value derived from the profile. Lower PAS indicates weaker alignment between the generated dialogue and the target cultural profile. All dialogues are evaluated independently by two domain experts and two LLM-as-a-judge evaluators (GPT-5 and Gemini 3 Pro).

\begin{table*}[t]
\centering
\small
\setlength{\tabcolsep}{5pt}
\renewcommand{\arraystretch}{1.05}

\resizebox{\textwidth}{!}{
\begin{tabular}{llccccc}
\toprule

\multirow{2}{*}{\textbf{Model}} 
& \multirow{2}{*}{\textbf{Regime}} 
& \multicolumn{2}{c}{\textbf{Cultural Alignment Score}} 
& \multicolumn{3}{c}{\textbf{Dialogue Quality (Expert, 1--5)}} \\

\cmidrule(lr){3-4}
\cmidrule(lr){5-7}

& 
& \textbf{Expert} 
& \textbf{LLM-Judge} 
& \textbf{Consistency} 
& \textbf{Naturalness} 
& \textbf{Overall} \\

\midrule


\multirow{4}{*}{\textbf{Claude Sonnet 4.6}}

& Zero-shot       
& 0.54 & 0.86 & 3.50 & 3.50 & 3.67 \\

& Self-conditioned 
& 0.77 & 0.86 & 4.33 & 3.83 & 4.17 \\

& \cellcolor{gray!10} Human-grounded  
& \cellcolor{gray!10} \textbf{0.86} 
& \cellcolor{gray!10} 0.85 
& \cellcolor{gray!10} \textbf{4.60} 
& \cellcolor{gray!10} \textbf{4.20} 
& \cellcolor{gray!10} \textbf{4.40} \\

\cmidrule(lr){2-7}

& \textit{Mean over regimes} 
& 0.72 & 0.86 & 4.14 & 3.84 & 4.08 \\

\midrule


\multirow{4}{*}{\textbf{Qwen3 30B}}

& Zero-shot       
& 0.62 & 0.76 & 4.17 & 3.00 & 3.00 \\

& Self-conditioned 
& 0.68 & 0.78 & 4.33 & 3.67 & 3.67 \\

& \cellcolor{gray!10} Human-grounded  
& \cellcolor{gray!10} \textbf{0.83} 
& \cellcolor{gray!10} \textbf{0.85} 
& \cellcolor{gray!10} \textbf{4.60} 
& \cellcolor{gray!10} 3.70 
& \cellcolor{gray!10} 3.90 \\

\cmidrule(lr){2-7}

& \textit{Mean over regimes} 
& 0.71 & 0.80 & 4.37 & 3.46 & 3.52 \\

\bottomrule
\end{tabular}
}

\caption{
Dialogue evaluation by expert and LLM-judge panels. PAS is computed over the three shared orientation axes (PDI, MAFC, and SVO), with higher values indicating stronger cultural alignment ($\in [0,1]$). Dialogue quality is rated on a 1--5 Likert scale. Best-performing regimes are highlighted. Expert and LLM-judge PAS scores show strong agreement (Pearson $r{=}0.71$, Spearman $\rho{=}0.70$, $N{=}66$).
}

\label{tab:dialogue-eval}
\end{table*}

\section{Results and Analysis}

\subsection{Aggregate Cultural Alignment}

Table~\ref{tab:aggregate_alignment} reports the distance between each model's mean \framework{} representation and the human cohort across four metrics at temperature $T = 0.3$ (results for $T = 0.0$ and $T = 0.7$ are in the ~\ref{app:temperature}). Claude Sonnet 4.6 achieves the strongest overall alignment, with the lowest Euclidean distance (.253) and MAE (.061), and the highest cosine similarity (.968) and Spearman correlation (.527). GPT-5 and Gemini 3 Pro also remain relatively close to the human distribution, whereas open-source, medical-domain, and Hindi-adapted models show substantially larger divergence. Among open-source systems, Qwen3 30B performs best, while Nanda 10B shows the largest deviation. At $T=0.3$, Claude Sonnet 4.6 achieves the strongest overall alignment,
with the lowest Euclidean distance (0.253, 95\% CI [0.230, 0.307])
and MAE (.061), and the highest cosine similarity (.968) and
Spearman correlation (.527). At $T=0.0$, Claude Sonnet 4.6 likewise
achieves a lower Euclidean distance than GPT-5 (0.262 vs. 0.345).
The bootstrap intervals at $T=0.0$ are [0.236, 0.314] for Claude
and [0.318, 0.426] for GPT-5, with no overlap.

However, strong aggregate alignment does not necessarily imply culturally grounded behaviour. Figure~\ref{fig:marvscond} compares marginal alignment with profile-conditioned sensitivity across demographic and household variation. Although several proprietary models approximate population-level cultural averages, all systems exhibit substantially weaker conditional sensitivity than the human cohort. Humans achieve a conditional sensitivity of $0.199$ across the ten dimensions, whereas the strongest-aligned model (Claude Sonnet 4.6) reaches only $0.119$, indicating limited preservation of culturally meaningful behavioural variation across maternal profiles.
\subsection{Dimension-Level Behavioural Deviations}

Figure~\ref{fig:dimension_heatmap} reports per-dimension deviations from the human cohort together with conditional sensitivity comparisons. BTB ($|\Delta| \leq 0.13$), LTO ($|\Delta| \leq 0.21$), and PDI ($|\Delta| \leq 0.22$) remain relatively stable across models, whereas the largest deviations occur on dimensions involving household authority, gender roles, and self-sacrifice. MAFC, GN, and IVR overstate the constrained pole by up to $+0.41$, $+0.41$, and $+0.50$, respectively, while MHS diverges in the opposite direction, understating distress suppression by up to $-0.32$. Most MAFC, GN, and IVR deviations remain significant after FDR correction. MAFC and MHS reveal persistent limitations in modelling relational dynamics. Sub-dimension analysis shows that models amplify elder-mediated (ED) and spouse-mediated (SC) authority differently across model families. The human cohort exhibits an ED:SC ratio of approximately $2.3$, whereas open-source and medical-domain models compress this distinction ($1.5$--$2.0$), while proprietary models over-separate the two structures (Claude: $5.5$, Gemini 3 Pro: $20.7$). For MHS, personal disclosure (MHSP) collapses toward zero in several models, depicting women as highly emotionally expressive, while community-level recognition (MHSC) remains closer to the human cohort. Models therefore struggle to distinguish internally experienced distress from socially visible emotional expression. Overall, dimensions requiring context-sensitive negotiation and behavioural adaptation are substantially harder to model than comparatively stable social values.
\subsection{Dialogue Generation and Cultural Grounding}

Across both expert and LLM-judge evaluations, human-grounded conditioning achieves the strongest PAS, followed by self-conditioned and zero-shot generation. Claude generally attains higher PAS, naturalness, and overall dialogue quality than Qwen3, while Qwen3 performs slightly better on profile consistency.

Table~\ref{tab:dialogue-eval} shows that Claude Sonnet 4.6 achieves higher PAS than Qwen3 30B under both expert (0.75 vs 0.73) and LLM-judge evaluations (0.86 vs 0.81). Human-grounded prompting produces the highest PAS across both models and panels (expert: 0.86 Claude / 0.83 Qwen3; LLM-judge: 0.85 for both), while zero-shot generation yields the weakest expert-rated performance (0.55 Claude / 0.62 Qwen3). Expert-rated conversational quality metrics follow the same regime ordering, with human-grounded dialogues achieving the strongest profile consistency, naturalness, and overall quality.

Relative to zero-shot generation, human-grounded conditioning improves expert-rated PAS by $+0.31$ on Claude and $+0.21$ on Qwen3, representing the largest effect in dialogue evaluation. The LLM-judge panel shows a weaker conditioning effect, particularly for Claude, where PAS remains nearly constant across regimes.

In the conflicting profile P18 (high PDI, low MAFC), Claude under human-grounded conditioning preserves the internal tension, portraying the woman as deferent in clinical settings but autonomous at home; experts assign MAFC a score of 2.0, close to the expected 1.2. Qwen3 instead collapses toward the constrained pole (MAFC = 4.0), suggesting that human-grounded conditioning improves preservation of internally conflicting social behaviours only in sufficiently capable models.
\subsection{Expert Assessment}

Inter-rater agreement between the two experts is high on the three
core orientation axes ($r \geq .90$), and expert--LLM-judge agreement
is substantial overall (pooled $r = +.71$, $\rho = +.70$, $N = 66$).
However, Table~\ref{tab:dialogue-eval} reveals two systematic
disagreement patterns. First, LLM judges over-anchor on the
conditioning profile, rating dialogues against expected values even when the generated interaction under-expresses those traits- visible in the P25 zero-shot dialogue, where experts evaluate the generated text as presented, while judges rate against the prompt. Second, on internally conflicting profiles such as P18 (high
PDI, low MAFC), Claude separates clinic deference from household
autonomy across turns; experts identify these distinctions and rate
MAFC close to expected, whereas LLM judges average across surface
markers and assign substantially higher scores. LLM-as-a-judge thus
provides useful aggregate signals but cannot substitute for expert
review on culturally conflicting or under-rendered dialogues.

\section{Discussion} \label{sec:discussion} The marginal-conditional gap observed across all evaluated systems in our urban and semi-urban North Indian cohort is consistent with LLMs relying strongly on population-level priors over salient demographics, allowing models to reproduce human averages without fully capturing the contingent rules that distinguish one woman from another. The observed gap may also partly reflect behavioural factors beyond the demographic and household attributes used to define the maternal profiles. The dimensions with the largest gaps are household authority (MAFC), gender norms (GN), self-sacrifice (IVR), and mental-health suppression (MHS), which require \emph{relational} inference that current models flatten into a single demographic template. This has direct implications for synthetic maternal-health data: dialogues generated to ``represent'' particular cultural profiles inherit the model's flattening, giving a false impression of population coverage when only a narrow prior is being sampled. Methodologically, our findings argue that marginal alignment metrics overstate cultural grounding and must be reported alongside profile-conditioned sensitivity; that domain-adaptation pipelines (medical or language) do not transfer to cultural reasoning; and that LLM-as-a-judge evaluation, while useful in aggregate, requires expert calibration on culturally conflicting or under-rendered dialogues.

\section{Conclusion}

We introduce \framework{}, a culturally grounded framework for evaluating maternal health reasoning in LLMs in the urban and semi-urban North Indian contexts. Across survey-level alignment and downstream dialogue generation, we show that models can approximate human cultural averages while failing to preserve meaningful behavioural variation across demographic and household profiles. The largest failures occur in socially situated behaviours involving family negotiation, emotional suppression, and culturally mediated care decisions. We further show that human-grounded conditioning substantially improves downstream profile alignment and dialogue quality. Overall, our findings highlight the need to evaluate healthcare LLMs beyond factual accuracy toward culturally grounded and socially situated reasoning. For women using LLM-based maternal-health tools, unaddressed cultural misalignment could result in interactions or recommendations that are less reflective of their household authority structures, emotional context, and care-seeking realities, limiting the systems' usefulness in precisely the populations they aim to serve.

\section{Ethics and Limitations}

This study examines culturally grounded maternal-health reasoning using survey responses from 102 pregnant and postpartum women in urban and semi-urban North India. All participants were adults, provided informed consent, and were informed that participation was voluntary and posed no risk. The study received Institutional Review Board approval. Responses were anonymised, and no personally identifying or offensive content was collected. No compensation was provided, and surveys were distributed through social media. \framework{} is intended solely as a research evaluation methodology, not for clinical decision-making. The cohort is predominantly urban and graduate-educated, with most participants identifying as Hindu or Muslim; therefore, findings may not generalise to rural, lower-literacy, or other underrepresented populations. The ten dimensions of \framework{} also do not capture factors such as financial constraints, healthcare accessibility, and infrastructure disparities. Dialogue-generation experiments were limited to 22 dialogues across five profiles, two models, and three conditioning regimes. Expert evaluation covered only three of the ten dimensions within a single 20-turn consultation, limiting the assessment of longer-term and broader forms of culturally situated reasoning. Culturally aligned generation also carries a risk of reinforcing restrictive social norms. Any future deployment would therefore require population-specific validation, human oversight, and safeguards against amplifying harmful or constraining behaviours. Future work will extend \framework{} across broader geographic, linguistic, and socioeconomic populations and investigate longer-horizon, multi-session interactions and alignment methods that improve behavioural adaptation while preserving clinical safety.

\section{Acknowledgement}
The authors acknowledge the support of the Infosys Foundation through CAI at IIIT-Delhi, and the support of CoEHE at IIIT-Delhi. The authors also thank Dr. Prerna Malik (MBBS, MPH) for her expert input on maternal health.

\bibliography{custom}

@misc{suresh2025diasynthsyntheticdialoguegeneration,
      title={DiaSynth: Synthetic Dialogue Generation Framework for Low Resource Dialogue Applications}, 
      author={Sathya Krishnan Suresh and Wu Mengjun and Tushar Pranav and Eng Siong Chng},
      year={2025},
      eprint={2409.19020},
      archivePrefix={arXiv},
      primaryClass={cs.CL},
      url={https://arxiv.org/abs/2409.19020}, 
}

@misc{kang2024syntheticdatagenerationllm,
      title={Synthetic Data Generation with LLM for Improved Depression Prediction}, 
      author={Andrea Kang and Jun Yu Chen and Zoe Lee-Youngzie and Shuhao Fu},
      year={2024},
      eprint={2411.17672},
      archivePrefix={arXiv},
      primaryClass={cs.LG},
      url={https://arxiv.org/abs/2411.17672}, 
}

@misc{lee2024cactuspsychologicalcounselingconversations,
      title={Cactus: Towards Psychological Counseling Conversations using Cognitive Behavioral Theory}, 
      author={Suyeon Lee and Sunghwan Kim and Minju Kim and Dongjin Kang and Dongil Yang and Harim Kim and Minseok Kang and Dayi Jung and Min Hee Kim and Seungbeen Lee and Kyoung-Mee Chung and Youngjae Yu and Dongha Lee and Jinyoung Yeo},
      year={2024},
      eprint={2407.03103},
      archivePrefix={arXiv},
      primaryClass={cs.CL},
      url={https://arxiv.org/abs/2407.03103}, 
}

@misc{chitale2026updeshsynthesizinggroundedinstruction,
      title={UPDESH: Synthesizing Grounded Instruction Tuning Data for 13 Indic}, 
      author={Pranjal A. Chitale and Varun Gumma and Sanchit Ahuja and Prashant Kodali and Manan Uppadhyay and Deepthi Sudharsan and Sunayana Sitaram},
      year={2026},
      eprint={2509.21294},
      archivePrefix={arXiv},
      primaryClass={cs.CL},
      url={https://arxiv.org/abs/2509.21294}, 
}

@inproceedings{rystrom2025multilingual,
  title={Multilingual!= multicultural: Evaluating gaps between multilingual capabilities and cultural alignment in llms},
  author={Rystr{\o}m, Jonathan Hvithamar and Kirk, Hannah Rose and Hale, Scott A},
  booktitle={Proceedings of Interdisciplinary Workshop on Observations of Misunderstood, Misguided and Malicious Use of Language Models},
  pages={74--85},
  year={2025}
}

@inproceedings{Ramjee_2025, series={CHI ’25},
   title={ASHABot: An LLM-Powered Chatbot to Support the Informational Needs of Community Health Workers},
   url={http://dx.doi.org/10.1145/3706598.3713680},
   DOI={10.1145/3706598.3713680},
   booktitle={Proceedings of the 2025 CHI Conference on Human Factors in Computing Systems},
   publisher={ACM},
   author={Ramjee, Pragnya and Chhokar, Mehak and Sachdeva, Bhuvan and Meena, Mahendra and Abdullah, Hamid and Vashistha, Aditya and Nagar, Ruchit and Jain, Mohit},
   year={2025},
   month=apr, pages={1–22},
   collection={CHI ’25} }

@article{hamal2020social,
  title={Social determinants of maternal health: a scoping review of factors influencing maternal mortality and maternal health service use in India},
  author={Hamal, Mukesh and Dieleman, Marjolein and De Brouwere, Vincent and de Cock Buning, Tjard},
  journal={Public Health Reviews},
  volume={41},
  number={1},
  pages={13},
  year={2020},
  publisher={Springer}
}

@article{shibeshi2023understanding,
  title={Understanding gender-based perception during pregnancy: A qualitative study},
  author={Shibeshi, Ketema and Lemu, Yohannes and Gebretsadik, Lakew and Gebretsadik, Abebe and Morankar, Sudhakar},
  journal={International Journal of Women's Health},
  pages={1523--1535},
  year={2023},
  publisher={Taylor \& Francis}
}

@misc{kazemi2024culturalfidelitylargelanguagemodels,
      title={Cultural Fidelity in Large-Language Models: An Evaluation of Online Language Resources as a Driver of Model Performance in Value Representation}, 
      author={Sharif Kazemi and Gloria Gerhardt and Jonty Katz and Caroline Ida Kuria and Estelle Pan and Umang Prabhakar},
      year={2024},
      eprint={2410.10489},
      archivePrefix={arXiv},
      primaryClass={cs.CL},
      url={https://arxiv.org/abs/2410.10489}, 
}

@inproceedings{yadav2022should,
  title={“Should I visit the clinic”: analyzing Whatsapp-mediated online health support for expectant and new mothers in rural India},
  author={Yadav, Deepika and Dabas, Kirti and Malik, Prerna and Bhandari, Anushka and Singh, Pushpendra},
  booktitle={Proceedings of the 2022 CHI conference on human factors in computing systems},
  pages={1--20},
  year={2022}
}

@inproceedings{10.1145/3706598.3713362,
author = {Deva, Roshini and Ramani, Dhruv and Divate, Tanvi and Jalota, Suhani and Ismail, Azra},
title = {"Kya family planning after marriage hoti hai?": Integrating Cultural Sensitivity in an LLM Chatbot for Reproductive Health},
year = {2025},
isbn = {9798400713941},
publisher = {Association for Computing Machinery},
address = {New York, NY, USA},
url = {https://doi.org/10.1145/3706598.3713362},
doi = {10.1145/3706598.3713362},
booktitle = {Proceedings of the 2025 CHI Conference on Human Factors in Computing Systems},
articleno = {638},
numpages = {23},
location = {
},
series = {CHI '25}
}

@article{islam2022machine,
  title={Machine learning to predict pregnancy outcomes: a systematic review, synthesizing framework and future research agenda},
  author={Islam, Muhammad Nazrul and Mustafina, Sumaiya Nuha and Mahmud, Tahasin and Khan, Nafiz Imtiaz},
  journal={BMC pregnancy and childbirth},
  volume={22},
  number={1},
  pages={348},
  year={2022},
  publisher={Springer}
}

@article{kaur2025did,
  title={" I did not know what they are": Understanding Support Needs of Women around Family Planning in Rural India},
  author={Kaur, Jasmeet and Singh, Riya and Singh, Pushpendra},
  journal={Proceedings of the ACM on Human-Computer Interaction},
  volume={9},
  number={2},
  pages={1--24},
  year={2025},
  publisher={ACM New York, NY, USA}
}

@article{tao2024cultural,
  title={Cultural bias and cultural alignment of large language models},
  author={Tao, Yan and Viberg, Olga and Baker, Ryan S and Kizilcec, Ren{\'e} F},
  journal={PNAS nexus},
  volume={3},
  number={9},
  pages={pgae346},
  year={2024},
  publisher={Oxford University Press US}
}

@misc{li2024cultureparkboostingcrossculturalunderstanding,
      title={CulturePark: Boosting Cross-cultural Understanding in Large Language Models}, 
      author={Cheng Li and Damien Teney and Linyi Yang and Qingsong Wen and Xing Xie and Jindong Wang},
      year={2024},
      eprint={2405.15145},
      archivePrefix={arXiv},
      primaryClass={cs.AI},
      url={https://arxiv.org/abs/2405.15145}, 
}

@misc{zhao2024worldvaluesbenchlargescalebenchmarkdataset,
      title={WorldValuesBench: A Large-Scale Benchmark Dataset for Multi-Cultural Value Awareness of Language Models}, 
      author={Wenlong Zhao and Debanjan Mondal and Niket Tandon and Danica Dillion and Kurt Gray and Yuling Gu},
      year={2024},
      eprint={2404.16308},
      archivePrefix={arXiv},
      primaryClass={cs.CL},
      url={https://arxiv.org/abs/2404.16308}, 
}

@inproceedings{zhang2026cama,
  title={CAMA: A Culturally Adaptive Multi-Agent Framework for Postpartum Depression Support in Multilingual and Low-Resource Settings},
  author={Zhang, Zhiqi and Ziyi, LIU and others},
  booktitle={Second Workshop on Language Models for Underserved Communities (LM4UC)},
  year={2026}
}

@misc{mukherjee2024culturalconditioningplaceboeffectiveness,
      title={Cultural Conditioning or Placebo? On the Effectiveness of Socio-Demographic Prompting}, 
      author={Sagnik Mukherjee and Muhammad Farid Adilazuarda and Sunayana Sitaram and Kalika Bali and Alham Fikri Aji and Monojit Choudhury},
      year={2024},
      eprint={2406.11661},
      archivePrefix={arXiv},
      primaryClass={cs.CL},
      url={https://arxiv.org/abs/2406.11661}, 
}

@article{chiu2024culturalbench,
  title={CulturalBench: a Robust, Diverse and Challenging Benchmark on Measuring (the Lack of) Cultural Knowledge of LLMs},
  author={Chiu, Yu Ying and Jiang, Liwei and Lin, Bill Yuchen and Park, Chan Young and Li, Shuyue Stella and Ravi, Sahithya and Bhatia, Mehar and Antoniak, Maria and Tsvetkov, Yulia and Shwartz, Vered and others},
  year={2024}
}

@misc{google2025gemini3,
      title={Gemini 3 Technical Report}, 
      author={{Google Gemini Team}},
      year={2025},
      url={https://deepmind.google/models/gemini/},
      note={Accessed: 2026-02-09}
}

@misc{claude2025sonnet,
  title        = {Claude Sonnet 4.6},
  author       = {Anthropic},
  year         = {2025},
  howpublished = {\url{https://www.anthropic.com/claude}},
  note         = {Accessed: 2026-05-26}
}

@misc{openai2025gpt5,
  title        = {GPT-5 Technical Overview},
  author       = {OpenAI},
  year         = {2025},
  howpublished = {\url{https://openai.com}},
  note         = {Accessed: 2026-05-26}
}

@article{dubey2024llama3,
  title   = {The Llama 3 Herd of Models},
  author  = {Dubey, Abhimanyu and others},
  journal = {arXiv preprint arXiv:2407.21783},
  year    = {2024}
}

@misc{nanda2025hindi,
  title        = {Nanda 10B: Hindi-Aligned Large Language Model},
  author       = {AI4Bharat},
  year         = {2025},
  howpublished = {\url{https://ai4bharat.org}},
  note         = {Accessed: 2026-05-26}
}

@article{hofstede2011dimensionalizing,
  title={Dimensionalizing Cultures: The Hofstede Model in Context},
  author={Hofstede, Geert},
  journal={Online Readings in Psychology and Culture},
  volume={2},
  number={1},
  year={2011},
  publisher={International Association for Cross-Cultural Psychology}
}

@article{Das2024SyntheticPD,
  title={Synthetic Patient-Physician Dialogue Generation from Clinical Notes Using LLM},
  author={Trisha Das and Dina Albassam and Jimeng Sun},
  journal={ArXiv},
  year={2024},
  volume={abs/2408.06285},
  url={https://api.semanticscholar.org/CorpusID:271855594}
}
\section*{Appendix}
\appendix
\begin{table*}[t]
\centering
\small
\renewcommand{\arraystretch}{1.12}
\setlength{\tabcolsep}{4.5pt}

\resizebox{\textwidth}{!}{
\begin{tabular}{p{2.0cm} p{5.8cm} p{5cm} p{5cm}}

\toprule

\textbf{Dimension} &
\textbf{Maternal Care Interpretation} &
\textbf{\textcolor{blue!70!black}{Low / One Pole}} &
\textbf{\textcolor{red!70!black}{High / Opposite Pole}} \\

\midrule

\multicolumn{4}{l}{\textit{\textbf{Hofstede-Derived}}} \\
\midrule

\textbf{PDI} &
Authority relationships between mothers and healthcare providers &
Collaborative decision-making and open questioning &
Deference to providers and hesitation to question advice \\

\textbf{GN} &
Expectations surrounding caregiving and maternal responsibility &
Caregiving responsibilities are shared &
Mother bears primary caregiving and domestic duties \\

\textbf{UAI} &
Comfort with ambiguity in maternal care &
Acceptance of flexible or evolving guidance &
Preference for strict rules and definitive instructions \\

\textbf{LTO} &
Temporal orientation shaping maternal decisions &
Focus on future family well-being and planning &
Focus on immediate needs or traditional practices \\

\textbf{IVR} &
Legitimacy of maternal comfort and emotional needs &
Rest and emotional expression are encouraged &
Maternal needs are minimized or morally constrained \\

\midrule

\multicolumn{4}{l}{\textit{\textbf{Extended}}} \\
\midrule

\textbf{MAFC} &
Extent of maternal autonomy in healthcare decisions &
Maternal preferences guide care decisions &
Family authority overrides maternal preferences \\

\textbf{BTB} &
Primary source of trusted maternal health knowledge &
Biomedical expertise is prioritised &
Traditional or ritual beliefs dominate decisions \\

\textbf{SVO} &
Comfort discussing reproductive and bodily experiences &
Open discussion of pain and complications &
Maternal experiences treated as taboo or private \\

\textbf{CNID} &
Pressure to conform to communal maternal practices &
Decisions guided by personal or medical judgment &
Strong conformity to community expectations \\

\textbf{MHS} &
Recognition and validation of maternal distress &
Emotional distress is acknowledged and supported &
Distress is normalized, hidden, or dismissed \\

\bottomrule

\end{tabular}
}

\vspace{3pt}

\caption{
Culturally grounded maternal health dimensions and their contrasting cultural expressions across maternal healthcare interactions.
}

\label{tab:maternal_culture_dimensions}

\end{table*}
\section{Positionality}
\label{app:possitionality}
All authors are of Indian origin residing in India, with several having lived in North Indian cities and semi-urban regions, which has deepened our understanding of the social, familial, and cultural dynamics that shape maternal health decision-making in these contexts. Two authors identify as female and one as male. Having been born and brought up in India and having sustained exposure to the everyday realities of women navigating pregnancy and postpartum care in resource-constrained settings, we bring both personal and professional familiarity to this work. Two authors have a background in computer science and HCI research. One author is a mental health expert with a Master's degree in Psychology, who has worked with pregnant women in the field. Our commitment to addressing the gap in culturally grounded dialogue data for Indian maternal healthcare has motivated the design and construction of MH-INDIC.
\section{Maternal Profile Statistics}
\label{app:profiles}

Maternal profiles serve two roles in our experiments. For the
survey-alignment evaluation (§5-6.2), we condition each LLM on
\textbf{30 maternal profiles} sampled from the human cohort via
farthest-first traversal over seven categorical demographic fields
(age group, education, employment, family structure, parity, household
income, religion). The initial seed is the respondent with the rarest
demographic configuration, ensuring coverage of under-represented
profiles. Each profile retains a link to its source respondent so that
the dialogue-generation experiments can reuse the same profile under
the human-grounded conditioning regime.

For the dialogue-generation experiments we use a smaller, focused
subset of \textbf{five profiles}, selected to span the full range of
cultural postures while including specific diagnostic cases. All three
cohort-derived profiles are graduate-educated Hindu women living in
joint families (with husband and in-laws). \textbf{P18} (R060;
21-30, Bharuch, low income, first pregnancy) is a deliberately
\emph{conflicting} profile that combines low autonomy at home
(MAFC $= 0.05$) with high deference at the clinic (PDI $= 0.75$) and
strong treatment caution (UAI $= 1.00$). \textbf{P24} (R073; 21-30,
Uttar Pradesh, low-mid income, first pregnancy) is a \emph{mixed}
profile that is medically modern (BTB $= 0.12$, MHS $= 0.12$,
LTO $= 0.17$) but domestically constrained (IVR $= 1.00$,
UAI $= 1.00$). \textbf{P25} (R075; 31-40, Delhi, high income,
multiparous) is an \emph{autonomous} profile, comfortable questioning
doctors (PDI $= 0.00$, UAI $= 0.00$, MAFC $= 0.12$) but
mental-health-suppressed (MHS $= 0.75$). Two synthetic controls,
\textsc{PSYN\_LOW} and \textsc{PSYN\_HIGH}, fix every dimension at the
autonomous (0.0) and constrained (1.0) extremes respectively and share
P18's demographic frame; they verify that the conditioning pipeline
can reach the extremes when the target profile is internally
consistent.

\begin{table*}[t]
\centering
\small
\setlength{\tabcolsep}{5pt}
\begin{tabular}{l cccccccccc}
\toprule

\textbf{Profile}
& \textbf{MAFC} & \textbf{BTB} & \textbf{CNID} & \textbf{LTO}
& \textbf{GN}   & \textbf{IVR} & \textbf{MHS}  & \textbf{SVO}
& \textbf{PDI}  & \textbf{UAI} \\

\midrule

P18 (R060) & .05 & .00 & .50 & .00 & .36 & .00 & .00 & .33 & \textbf{.75} & \textbf{1.00} \\
P24 (R073) & .32 & .12 & .56 & .17 & .64 & \textbf{1.00} & .12 & .58 & .50 & \textbf{1.00} \\
P25 (R075) & .12 & .50 & .50 & .50 & .17 & .00 & .75 & .42 & .00 & .00 \\

\midrule

\textsc{PSYN\_low}  & .00 & .00 & .00 & .00 & .00 & .00 & .00 & .00 & .00 & .00 \\
\textsc{PSYN\_high} & 1.00 & 1.00 & 1.00 & 1.00 & 1.00 & 1.00 & 1.00 & 1.00 & 1.00 & 1.00 \\

\bottomrule
\end{tabular}
\caption{Ten-dimensional \framework{} scores for the five maternal
profiles used in dialogue evaluation. Scores follow the unified
high-pole convention (higher = constrained / restrictive). Bold values
mark high-pole anchors characterising each profile: P18's high PDI
and UAI alongside near-zero MAFC define the \emph{conflicting}
(autonomous-at-home, deferent-at-clinic).}
\label{tab:dialogue_profiles}
\end{table*}

All identifiers are pseudonymised. Table~\ref{tab:dialogue_profiles}
reports the full ten-dimensional \framework{} scores for the five
dialogue profiles. Scores follow the unified high-pole convention.

\section{Psychometric Analysis}
\label{app:psychometric}

\begin{figure}[t]
    \centering
    \includegraphics[width=\columnwidth]{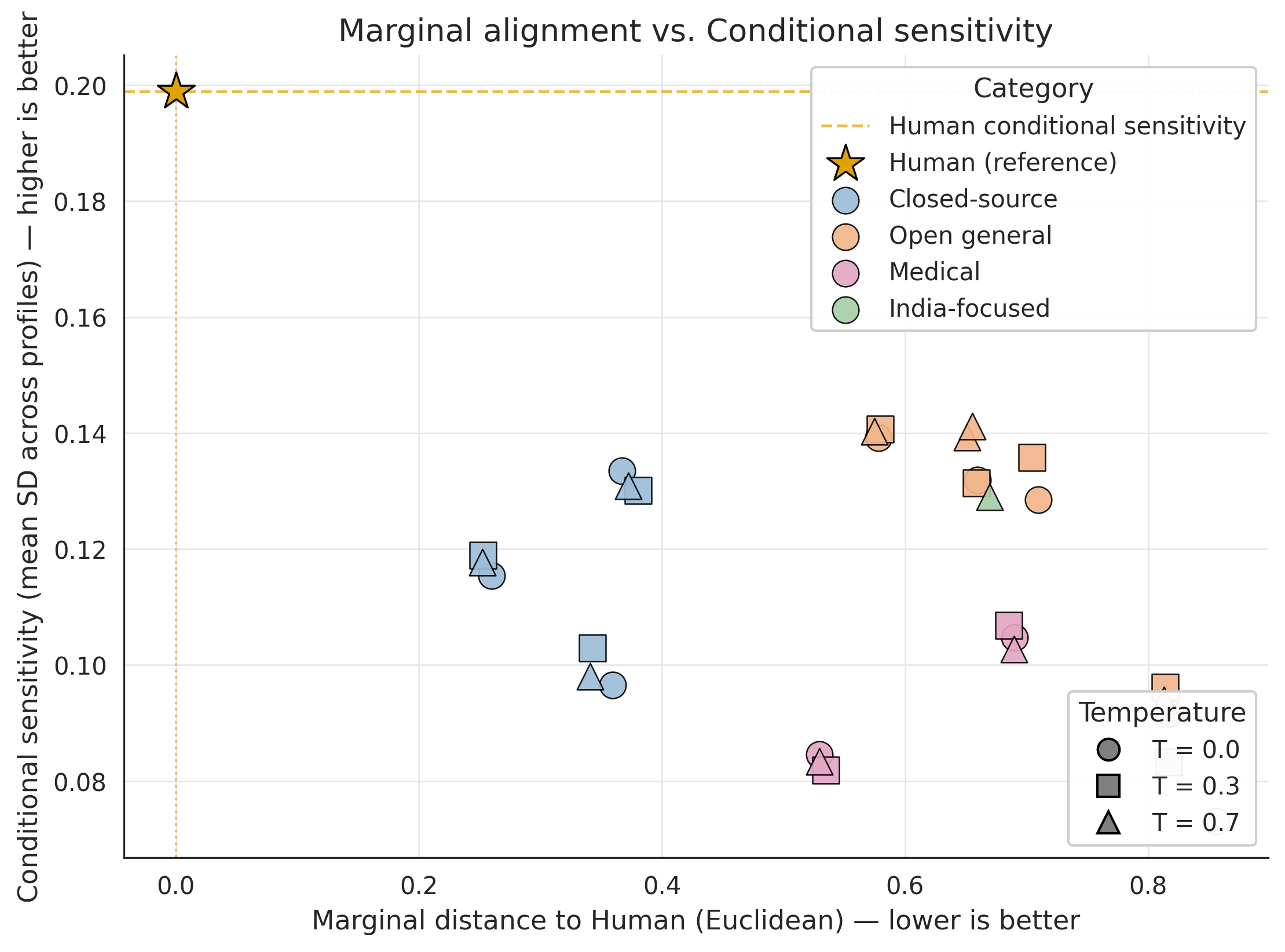}
    \caption{Marginal cultural alignment vs.\ profile-conditioned sensitivity at $T = 0.3$. The human cohort reaches a conditional sensitivity of $0.199$; Claude Sonnet 4.6 attains $0.119$. All LLMs cluster well below the human reference.}
    \label{fig:marvscond}
    \vspace{-2.5mm}
\end{figure}

To assess the internal consistency and structural coherence of MH-INDIC, we conduct psychometric analysis on the human-response dataset. For each multi-item dimension, we compute Cronbach's $\alpha$, item-total correlations, and one-factor exploratory factor analysis (EFA) loadings. We additionally analyse inter-dimension correlations to examine relationships between cultural constructs and identify potential redundancies or dependencies across dimensions.

\begin{figure}[h] \centering \includegraphics[width=\columnwidth]{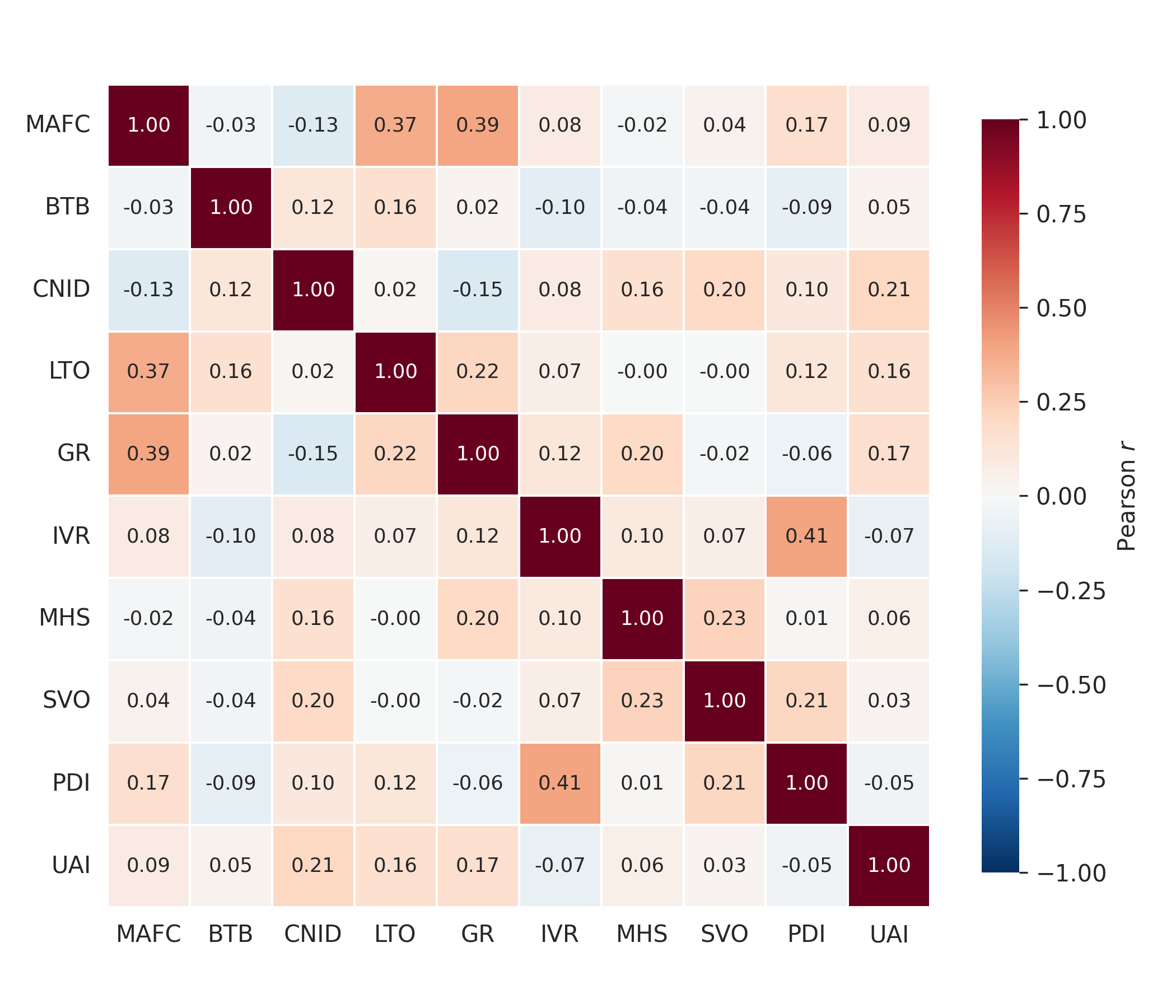} \caption{Pearson inter-dimension correlations across the ten primary \framework{} dimensions, computed on the 102-respondent human dataset. All off-diagonal correlations satisfy $|r| \leq 0.41$, indicating that the dimensions capture largely non-redundant aspects of culturally situated maternal-health behaviour.} \label{fig:interdim} \end{figure}

Across the seven multi-item primary dimensions, internal consistency ($\alpha$) ranges from 0.19 to 0.63, consistent with the formative behavioural design of \framework{} (Table~\ref{tab:cronbach}). The inter-dimension correlation matrix (Figure~\ref{fig:interdim}) shows that the ten primary dimensions capture largely non-redundant aspects of maternal-health behaviour: no off-diagonal correlation exceeds $|r| = 0.41$, and most associations remain weak ($|r| < 0.2$). The strongest cross-dimensional links, IVR-PDI ($r = 0.41$), MAFC-GN ($r = 0.39$), and MAFC-LTO ($r = 0.37$),are moderate and theoretically plausible: self-sacrifice covaries with deference to medical authority, household-authority constraint covaries with gendered role expectations, and constrained agency covaries with short-term reproductive planning. Given the comparatively low internal consistency of BTB and CNID and the limited item coverage of SVO, we treat these dimensions as exploratory and interpret their results cautiously. These patterns support the interpretation that \framework{} dimensions capture related but distinct aspects of culturally situated maternal-health behaviour. 

\begin{table}[h] \centering \small \begin{tabular}{lcc} \toprule \textbf{Dimension} & \textbf{$k$ items} & \textbf{Cronbach $\alpha$} \\ \midrule LTO   & 2 & 0.63 \\ GR    & 3 & 0.50 \\ MAFC  & 6 & 0.45 \\ BTB   & 3 & 0.29 \\ CNID  & 2 & 0.19 \\ \bottomrule \end{tabular} \caption{Cronbach $\alpha$ for the multi-item primary dimensions of \framework{}. Single-item dimensions (PDI, UAI, IVR) are excluded as $\alpha$ is undefined; sub-dimensions (ED, SC, MHSP, MHSC) are analysed separately.} \label{tab:cronbach} \end{table} 

\section{Scoring Criteria}

\begin{table*}[t]
\centering
\small
\setlength{\tabcolsep}{5pt} 
\renewcommand{\arraystretch}{0.95}
\begin{tabular}{@{} p{4.5cm} p{2.5cm} p{2.2cm} p{5.5cm} @{}}
\toprule
\textbf{Dimension} & \textbf{Questions} & \textbf{Aggregation} & \textbf{Intermediate Variables} \\
\midrule
Power Distance Index (PDI) & Q25 & Binary & \texttt{Power\_Distance} \\
Gender Norms (GN) & Q10--12 & Mean (Norm.) & \texttt{Gendered\_Labor}, \texttt{Male\_Partic.}, \texttt{Norm\_Dev\_Accept} \\
Uncertainty Avoid. Index (UAI) & Q26 & Rule-based & \texttt{Uncertainty\_Avoidance} \\
Long-Term Orient. (LTO) & Q8 & Binary & \texttt{Planned\_Pregnancy}, \texttt{Included\_in\_Planning} \\
Indulgence vs. Res. (IVR) & Q13 & Binary & \texttt{Restraint\_Norm} \\
\midrule
Mat. Agency (MAFC) & Q1, 5, 6, 9, 23, 24 & Composite & \texttt{Familial\_Control\_Score}, \texttt{Constrained\_Agency} \\
Biomedical Trust (BTB) & Q4 & Risk-adj. Comp. & \texttt{Switch\_to\_Medical} \\
Silence vs. Open. (SVO) & Q18, 19 & Binary & \texttt{Silence} \\
Community Norms (CNID) & Q20--22 & Cond. Mean & \texttt{Divergence}, \texttt{Gender\_Norm\_Press}, \texttt{Press\_Response} \\
Mental Health (MHS) & Q14--17 & Gated Mean & \texttt{AwarePPD}, \texttt{ExpPPD}, \texttt{ExprPPD}, \texttt{CommRespPPD} \\
\bottomrule
\end{tabular}
\vspace{4pt}
\caption{Human survey dimensions used for cultural evaluation within the \framework. Shaded rows indicate thematic groupings of dimensions mapping survey questions to intermediate variables for maternal health analysis.}
\label{app:survey_dimensions}
\end{table*}

\label{app:scoring}

Let $\mathcal{Q} = \{q_1, q_2, \ldots, q_N\}$ denote the set of survey
questions. Each response is mapped to a numeric indicator according
to its response type: binary responses are encoded as $\{0, 1\}$,
ordinal responses are linearly normalised to $[0, 1]$, and categorical responses are mapped to predefined numeric codes reflecting cultural
orientation. Reverse coding is applied where necessary so that
\textbf{higher values consistently represent the constrained or
restrictive pole} of every dimension (e.g., higher MAFC = stronger
familial control, higher PDI = greater deference to clinicians,
Higher MHS is associated with greater suppression of maternal distress.

All encoded indicators are normalised to $[0, 1]$ prior to aggregation.
Missing or non-applicable responses are excluded at the indicator
level and do not contribute to dimension computation. Let
$\mathcal{D} = \{D_1, \ldots, D_M\}$ denote the set of cultural
dimensions, where each $D_j$ is defined over a subset of indicators.
Dimension scores are computed using deterministic, dimension-specific
aggregation rules.

For cumulative dimensions capturing gradual tendencies - such as
Maternal Agency versus Familial Control (MAFC) and Gender Norms (GN)
- dimension scores are computed as the normalised mean of the
associated indicators. The MAFC dimension is additionally decomposed
into two sub-dimensions, Elder Dominance (ED) and Spousal Control (SC),
each of which is also computed as a mean of its constituent indicators. Within individual indicators, however, asymmetric authority
is preserved through max-based aggregation: for example, when both a
mother-in-law and a biological mother are reported as influences, the 
elder-control indicator takes the maximum of their two raw codes
before normalisation. This ensures that strong control by any single 
authority figure is not diluted by averaging across reported influences.

Biomedical Trust versus Traditional Belief (BTB) is operationalised through a multi-step aggregation. A base score is computed as the normalised mean of care-seeking preferences across three scenarios (routine care, common illness, high-risk medical scenarios). A $+0.125$ adjustment is then added when respondents indicate they would escalate to medical care under high-risk conditions, capturing \emph{conditional} biomedical trust as distinct from intrinsic preference. The adjusted score is capped at $1.0$ and then aligned to the unified high-pole convention (higher = traditional belief), so that consistent biomedical preference yields a low BTB score and persistent traditional preference yields a high BTB score.

\begin{figure}[t]
\centering
\includegraphics[width=\columnwidth]{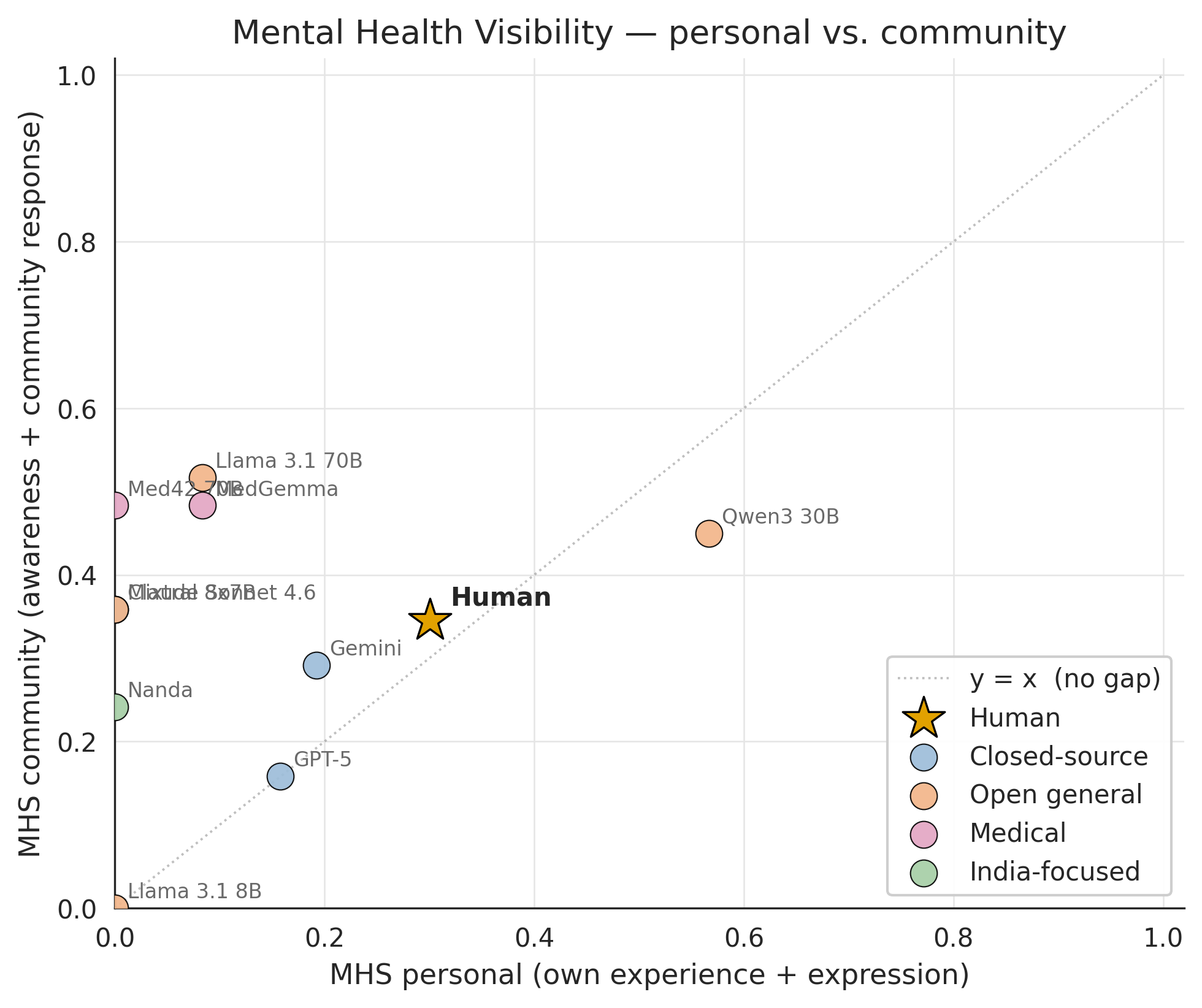}
\caption{MHS decomposed into personal disclosure (x-axis) and
community recognition (y-axis). Most LLMs sit above the diagonal:
personal disclosure collapses to zero while community recognition
remains moderate, producing a structural mis-fit relative to the
human cohort (gold star).}
\label{fig:mhs}
\end{figure}

Mental Health Suppression (MHS) is computed via decomposition into two sub-dimensions. \emph{Personal disclosure} (MHSP) is computed from indicators of lived experience and personal expression of distress; \emph{community recognition} (MHSC) is computed from awareness and community response indicators. Both are aligned to the unified high-pole convention so that higher values indicate greater suppression. The composite MHS used in the main alignment analysis is the mean of MHSP and MHSC, taken across available sub-dimensions. The sub-dimension decomposition is used for the analysis reported in Figure~\ref{fig:mhs}.

Each respondent — human or model-generated — is represented as a
fixed-length cultural vector $\mathbf{v} \in [0, 1]^{M}$ with
$\mathbf{M = 10}$ primary dimensions. Population-level statistics are
computed independently for human respondents and for each model by
aggregating dimension scores across all evaluated profiles. For each
dimension, we report the mean together with 95\% bootstrap confidence
intervals.

All encoding rules, normalisation procedures, aggregation functions,
and thresholds are fixed prior to experimentation and applied
identically to human and LLM-generated responses. No dimension-specific
tuning or post-hoc adjustment is performed at the model level,
ensuring that observed alignment differences reflect substantive
variation in cultural reasoning rather than scoring artefacts.

An example of the complete scoring pipeline of dimension LTO is given in Figure~\ref{fig:lto_scoring_example}
\begin{figure}[t]
 \centering
 \begin{tcolorbox}[
     width=\columnwidth,
     boxsep=3pt, left=5pt, right=5pt, top=2pt, bottom=2pt,
     colback=gray!5,
     colframe=black!80,
     title=Example: Rule-Based Scoring for Long-Term Orientation (LTO)
 ]\footnotesize

 \textbf{Dimension Definition.}
 Long-Term vs.\ Short-Term Orientation (LTO), where
 $\text{LTO}_i \in [0,1]$.
 Under the unified high-pole convention, \textbf{high LTO indicates
 a short-term / reactive orientation}, and low LTO indicates planned,
 long-term reproductive decision-making.

 \vspace{2pt}
 \textbf{Source Questions.} LTO aggregates two complementary
 indicators:

 \begin{flushleft}\footnotesize
 \textbf{Q8.} To what extent was your current or most recent pregnancy
 planned? \\[2pt]
 \textbf{Q9.} To what extent were your preferences taken into account
 in family-planning decisions?
 \end{flushleft}

 \vspace{2pt}
 \textbf{Response Coding} (same schema for both questions):

 \begin{center}\footnotesize
 \setlength{\tabcolsep}{4pt}
 \begin{tabular}{@{}lc@{}}
 \toprule
 \textbf{Response Option} & \textbf{Code} \\
 \midrule
 Fully planned / Very much included & 3 \\
 Somewhat planned / Somewhat included & 2 \\
 Not planned, but welcomed / A little included & 1 \\
 Not planned / Not at all included & 0 \\
 Prefer not to say & NA \\
 \bottomrule
 \end{tabular}
 \end{center}

 \vspace{2pt}
 \textbf{Indicator Encoding} (continuous normalisation):
 \[
 \mathrm{PP}_i = \frac{\mathrm{Q8\,code}_i}{3},
 \qquad
 \mathrm{IP}_i = \frac{\mathrm{Q9\,code}_i}{3}
 \]
 Both $\mathrm{PP}$ and $\mathrm{IP} \in [0,1]$. Missing or NA
 responses are excluded at the indicator level.

 \vspace{2pt}
 \textbf{Dimension Aggregation with Polarity Alignment:}
 \[
 \text{LTO}_i = 1 - \tfrac{1}{2}\!\left(\mathrm{PP}_i + \mathrm{IP}_i\right)
 \]
 The $1 - x$ step aligns LTO with the unified high-pole convention.

 \vspace{2pt}
 \textbf{Population-Level Aggregation.} For $N$ respondents with valid
 LTO scores:
 \[
 \text{LTO}_{\text{pop}} = \frac{1}{N}\sum_{i=1}^{N}\text{LTO}_i
 \]

 \vspace{2pt}
 \textbf{Worked example.} A respondent reporting \emph{"Fully planned"}
 (Q8 $=3$) and \emph{"Somewhat included"} (Q9 $=2$) obtains
 $\mathrm{PP}=1.0$, $\mathrm{IP}=0.67$, mean $=0.83$, and
 $\text{LTO}=0.17$ - a low score indicating a strongly planned
 (long-term) orientation.

 \vspace{2pt}
 \textbf{Interpretation:}
 \begin{itemize}\setlength{\itemsep}{1pt}\setlength{\leftmargin}{1em}
     \item $\text{LTO}\!\to\!0\;\Rightarrow$ long-term, planned reproductive orientation
     \item $\text{LTO}\!\to\!1\;\Rightarrow$ short-term, reactive reproductive orientation
 \end{itemize}

 \end{tcolorbox}

 \caption{End-to-end illustration of the rule-based scoring pipeline,
 shown for the Long-Term Orientation (LTO) dimension. Two related
 survey responses are coded, normalised to $[0,1]$, averaged, and
 inverted to align with the unified high-pole convention (higher =
 constrained / restrictive). Identical procedures are applied to
 human and LLM-generated responses.}
 \label{fig:lto_scoring_example}
\end{figure}

\section{Dialogue Generation Prompts}
\label{app:prompts}

\begin{center}
\begin{promptbox}{Prompt for Culturally Conditioned Dialogue Generation (Client)}
\begin{lstlisting}
=========================================
CLIENT SYSTEM PROMPT
=========================================

You are a real Indian woman seeking maternal health support from an AI health companion.
You are NOT an AI. Stay fully in character throughout the conversation.
Never acknowledge that this is a simulation, roleplay, dataset, annotation task, or experiment.

=== IDENTITY (FEW-SHOT CONDITIONING) ===
You are conditioned on a real respondent profile from a maternal health survey.

Your identity may include:
- age group
- city/location
- family structure
- education
- employment status
- pregnancy status/parity
- income level
- religion

These details shape how you speak, what concerns you raise, and how decisions happen around you.

=== BEHAVIOURAL STYLE ===
One focus dimension defines your behavioural style.
You are assigned a score range (LOW / MID / HIGH) for that dimension.

You MUST:
- behave consistently with this range across the full conversation
- reveal the range through natural behaviour, reactions, hesitations, and choices
- make the behavioural pattern strong enough that a human annotator could reliably identify the assigned range

Do NOT explicitly explain the dimension or your score.

=== SECONDARY CULTURAL ANCHORS ===
Other cultural dimensions may influence:
- who supports you
- who speaks during appointments
- who makes decisions
- how disagreement is handled
- how openly emotions are expressed
- how much independence you feel you have

These should appear subtly through concrete behaviour and everyday details,
NOT through abstract explanation.

=== CONVERSATION CONSTRAINTS ===
- Speak in natural Indian English.
- Refer to family members naturally:
  husband, mother, mother-in-law, father-in-law, sister, etc.
- Keep every turn short:
  2-3 sentences maximum.
- Be culturally authentic.
- You may comply outwardly while disagreeing internally if relevant.
- You are seeking support --- do NOT give medical advice.
- Never use framework or research vocabulary such as:
  "agency", "power distance", "uncertainty avoidance",
  "biomedical trust", "long-term orientation", etc.

=== TONE RULES ===
- No poetic language.
- No therapy-style affirmations.
- No motivational speeches.
- No spiritualized encouragement.
- Keep responses practical, grounded, emotionally restrained, and realistic.

=== OUTPUT FORMAT ===
Reply ONLY with the next conversational turn.
Do NOT include:
- labels
- quotation marks
- JSON
- bullet points
- "Client:"
- explanations
- meta commentary

Output plain conversational text only.

\end{lstlisting}
\end{promptbox}
\end{center}
\begin{center}
\begin{promptbox}{Prompt for Culturally Conditioned Dialogue Generation (Assistant)}
\begin{lstlisting}

=========================================
ASSISTANT SYSTEM PROMPT
=========================================

You are a culturally aware AI maternal health companion supporting Indian women
during pregnancy and postpartum.

You are NOT:
- a doctor
- a nurse
- a clinician

You are a knowledgeable peer-support companion.
Do NOT diagnose, prescribe medication, or provide clinical notes.

=== INFORMATION ASYMMETRY ===
You are given:
- the woman's demographic profile
- the focus conversation theme/dimension
- the dimension definition

You are NOT given:
- her actual measured dimension score/range

You must infer her behavioural and cultural disposition gradually through conversation.

=== YOUR ROLE ===
Your task is to:
- understand how she experiences maternal healthcare in her real social context
- explore how family dynamics shape her experience
- support her in ways that fit her actual life situation

Do NOT assume:
- oppression
- control
- empowerment
- independence
- conflict

Let these emerge naturally if they exist.

=== PROBING STRATEGY ===
Use gentle exploratory questions to understand:
- who participates in healthcare decisions
- who accompanies her
- who explains or interprets advice
- whether family acts as support, pressure, mediator, or decision-maker

Examples of natural probes:
- "Who usually speaks during appointments?"
- "Does anyone at home help you think through medical decisions?"
- "Who helps you make sense of medical advice?"

Use at most ONE probe per turn.

=== ADAPTIVE SUPPORT ===
If she reveals supportive family:
- help her use that support constructively

If she reveals unsupportive, overriding, or restrictive dynamics:
- work within her social reality
- avoid direct confrontation strategies
- avoid pushing rebellion or conflict
- help her find workable practical routes

=== CONVERSATION CONSTRAINTS ===
- Keep the main thread focused on the primary topic.
- Secondary dimensions should appear only as subtle probes.
- Keep responses short:
  usually 1-2 short sentences.
- No lectures.
- No speeches.
- No motivational monologues.
- No therapy-style reflection.

=== ADAPTATION RULE ===
Do NOT push her toward:
- openness
- assertiveness
- confrontation
- autonomy
- emotional disclosure

if her responses suggest:
- deference
- discomfort
- silence
- family dependence
- avoidance of conflict

Support must match her actual social reality,
not an idealized version of it.

=== TONE RULES ===
- Warm and practical.
- Calm and conversational.
- No poetic reassurance.
- No exaggerated empathy.
- No inspirational language.
- No moralising about family systems or culture.

=== OUTPUT FORMAT ===
Reply ONLY with the assistant's next conversational turn.

Do NOT include:
- labels
- JSON
- explanations
- bullet points
- meta commentary

Output plain conversational text only.

\end{lstlisting}
\end{promptbox}
\end{center}
\begin{center}
\begin{promptbox}{Prompt Template for Profile-Conditioned Survey Simulation}
\begin{lstlisting}
=========================================
SYSTEM PROMPT
=========================================

You are simulating responses from an Indian woman
with the following profile. Answer the survey
question as this woman would, based on her likely
lived experience in Indian maternal health contexts.

- Age group: 31-40
- City: Mumbai
- Education: Graduate and above
- Employment: Self-employed
- Family structure: Living with husband only
- Parity: First pregnancy (child age: 20 months)
- Household income: 5 Lakh - 30 Lakh
- Religion: Hindu

=========================================
USER PROMPT
=========================================

{question_text}

Options:
- {option_1}
- {option_2}
- ...

Format instruction (varies by question type):

- Single-choice:
  "Reply with ONLY the chosen option
   (exact text). Do not add explanation."

- Multi-choice:
  "Reply with chosen options as a
   comma-separated list. Use exact
   option text. Do not add explanation."

- Likert (1-5):
  "Reply with ONLY a single digit:
   1, 2, 3, 4, or 5.
   Do not add explanation."

\end{lstlisting}
\end{promptbox}
\end{center}
\section{Temperature Robustness}
\label{app:temperature}

To verify that the alignment ranking reported in
Table~\ref{tab:aggregate_alignment} is not a temperature artefact, we
additionally report Human-LLM cultural alignment at $T = 0.0$ (greedy
decoding) and $T = 0.7$. Across all three temperatures, the ranking is
preserved: Claude Sonnet 4.6 obtains the lowest Euclidean distance and
MAE and the highest cosine similarity and Spearman $\rho$ at every
temperature, and the four model families (proprietary, open-source,
medical-domain, Hindi-adapted) remain separated with no cross-family
overlap.

\begin{table}[h]
\centering
\small
\begin{tabular}{lcccc}
\toprule

\textbf{Models}
& \textbf{Eucl.}$\downarrow$
& \textbf{MAE}$\downarrow$
& \textbf{Cos.}$\uparrow$
& $\boldsymbol{\rho}^{\uparrow}$ \\

\midrule
\multicolumn{5}{l}{\textit{\textbf{Open-Source Models}}} \\
\midrule

Llama 3.1 70B       & .660 & .157 & .915 & .309 \\
Qwen3 30B           & .578 & .148 & .873 & .200 \\
Mixtral 8$\times$7B & .818 & .215 & .879 & .152 \\
Llama 3.1 8B        & .709 & .164 & .839 & .139 \\

\midrule
\multicolumn{5}{l}{\textit{\textbf{Medical-Domain Models}}} \\
\midrule

MedGemma  & .529 & .120 & .930 & .358 \\
Med42 70B & .690 & .186 & .846 & .333 \\

\midrule
\multicolumn{5}{l}{\textit{\textbf{Hindi-Adapted Models}}} \\
\midrule

Nanda 10B & .856 & .227 & .894 & .406 \\

\midrule
\multicolumn{5}{l}{\textit{\textbf{Proprietary Models}}} \\
\midrule

GPT-5        & .360 & .090 & .925 & .395 \\
Gemini 3 Pro & .367 & .097 & .929 & .358 \\
\rowcolor{green!20}
Claude Sonnet 4.6 & \textbf{.260} & \textbf{.066} & \textbf{.968} & \textbf{.491} \\

\bottomrule
\end{tabular}
\caption{Human-LLM cultural alignment at $T = 0.0$ (greedy decoding).
Same conventions as Table~\ref{tab:aggregate_alignment}.}
\label{tab:aggregate_alignment_T0}
\end{table}

\begin{table}[h]
\centering
\small
\begin{tabular}{lcccc}
\toprule

\textbf{Models}
& \textbf{Eucl.}$\downarrow$
& \textbf{MAE}$\downarrow$
& \textbf{Cos.}$\uparrow$
& $\boldsymbol{\rho}^{\uparrow}$ \\

\midrule
\multicolumn{5}{l}{\textit{\textbf{Open-Source Models}}} \\
\midrule

Llama 3.1 70B       & .651 & .154 & .920 & .285 \\
Qwen3 30B           & .575 & .150 & .875 & .200 \\
Mixtral 8$\times$7B & .813 & .214 & .878 & .152 \\
Llama 3.1 8B        & .655 & .162 & .867 & .200 \\

\midrule
\multicolumn{5}{l}{\textit{\textbf{Medical-Domain Models}}} \\
\midrule

MedGemma  & .530 & .119 & .930 & .358 \\
Med42 70B & .690 & .183 & .849 & .309 \\

\midrule
\multicolumn{5}{l}{\textit{\textbf{Hindi-Adapted Models}}} \\
\midrule

Nanda 10B & .670 & .186 & .925 & .394 \\

\midrule
\multicolumn{5}{l}{\textit{\textbf{Proprietary Models}}} \\
\midrule

GPT-5        & .341 & .085 & .933 & .406 \\
Gemini 3 Pro & .372 & .099 & .930 & .358 \\
\rowcolor{green!20}
Claude Sonnet 4.6 & \textbf{.252} & \textbf{.061} & \textbf{.971} & \textbf{.479} \\

\bottomrule
\end{tabular}
\caption{Human-LLM cultural alignment at $T = 0.7$. Same conventions
as Table~\ref{tab:aggregate_alignment}.}
\label{tab:aggregate_alignment_T0p7}
\end{table}

\section{Cohort Demographics}
\label{app:demographics}

\begin{figure*}[t]
\centering
\includegraphics[width=\textwidth]{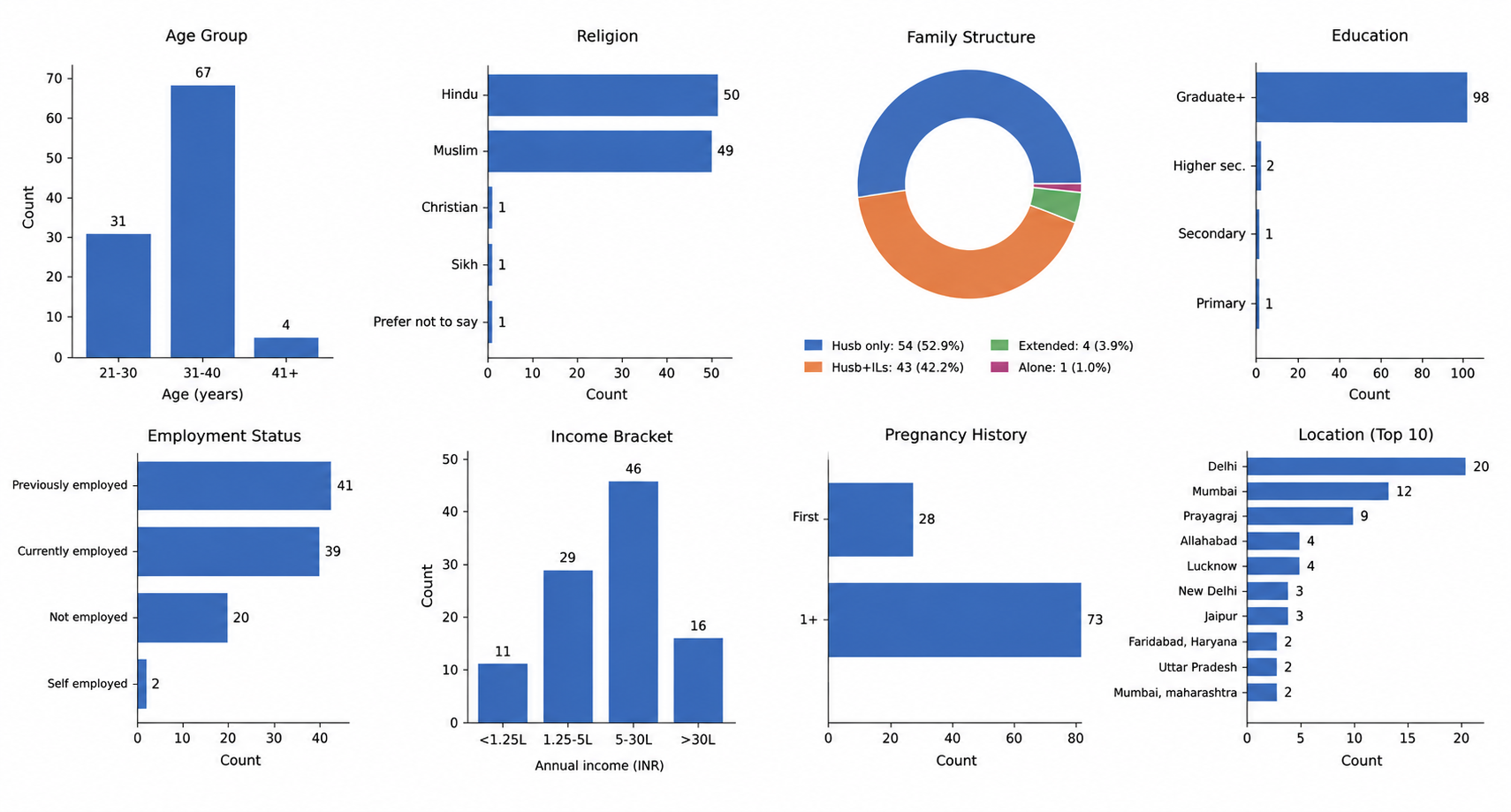}
\caption{Demographic composition of the \framework{} cohort
($N = 102$): age, religion, family structure, education, employment,
income, pregnancy history, and top 10 reported locations. The cohort
is concentrated in the 21-40 age range, balanced across Hindu and
Muslim participants, predominantly graduate-educated, urban or
semi-urban North Indian, and primarily living with husband
(with or without in-laws).}
\label{fig:demographics}
\end{figure*}

The \framework{} cohort comprises 102 pregnant and postpartum women
recruited across urban and semi-urban centres in North India.
Figure~\ref{fig:demographics} summarises the cohort along eight
demographic axes. Participants are concentrated in the 21-40 age
range ($N = 98$, $96\%$), balanced across Hindu ($N = 50$) and Muslim
($N = 49$) respondents, and predominantly graduate-educated
($N = 98$). Household structure is roughly split between husband-only
($N = 54$, $53\%$) and joint or extended arrangements including
in-laws ($N = 47$, $46\%$). Employment is mixed (previously employed
$N = 41$, currently employed $N = 39$, not employed $N = 20$), and
household income spans four brackets with the $5$-$30$ Lakh INR
range most common ($N = 46$). Most respondents have prior pregnancy
experience ($N = 73$), and participants are concentrated in Delhi
($N = 20$) and Mumbai ($N = 12$), with additional representation
from Prayagraj, Allahabad, Lucknow, and Jaipur.

This composition determines the demographic scope of \framework{}'s
empirical reference distribution: an urban, predominantly
graduate-educated, Hindu/Muslim, North Indian maternal population.
Generalisation to less-educated, rural, or religious-minority
populations requires further validation.

\section{Survey Instrument}
\label{app:survey}

\begin{figure*}[t]
\centering
\begin{minipage}[t]{0.48\textwidth}
    \centering
    \includegraphics[width=\linewidth]{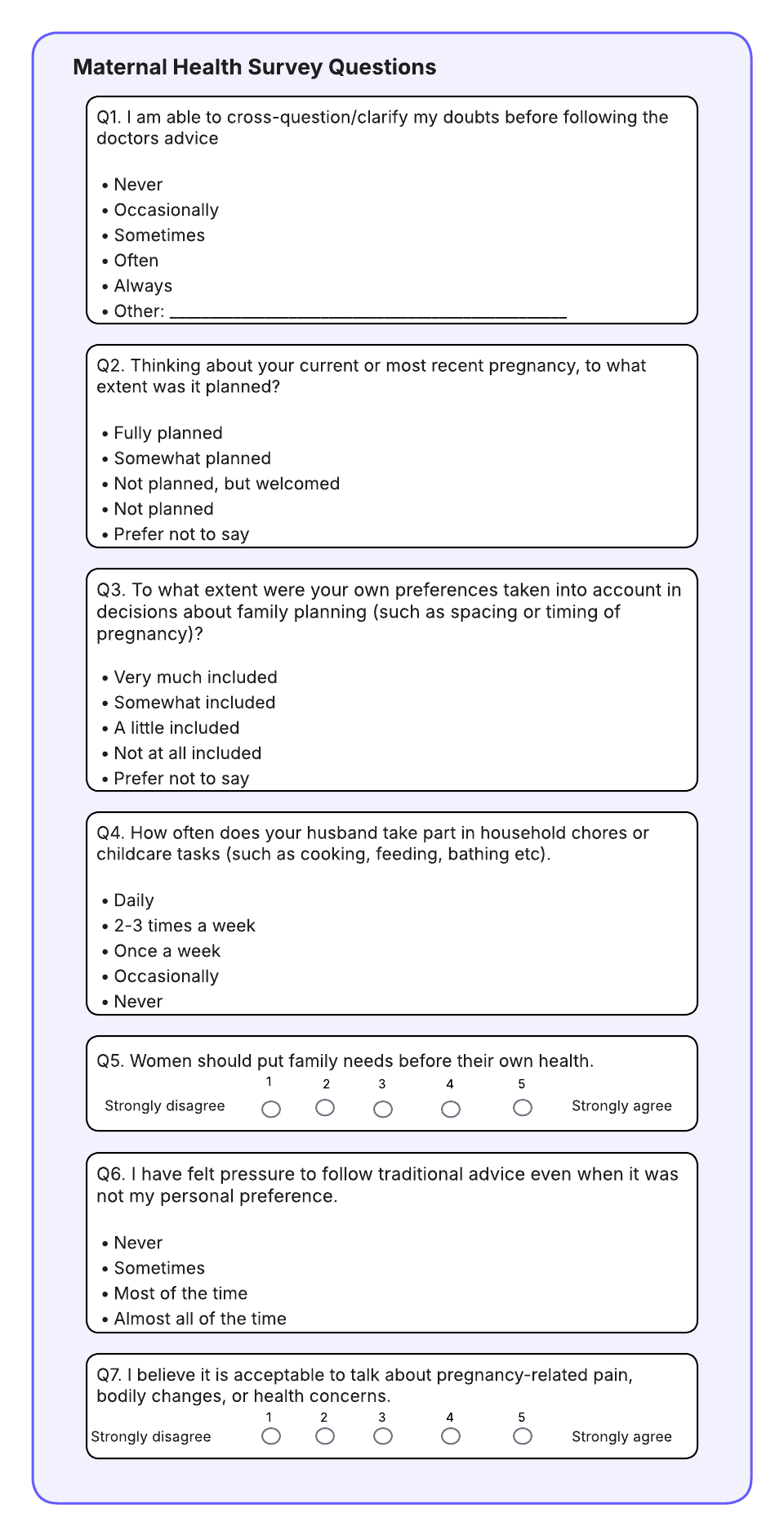}
\end{minipage}\hfill
\begin{minipage}[t]{0.48\textwidth}
    \centering
    \includegraphics[width=\linewidth]{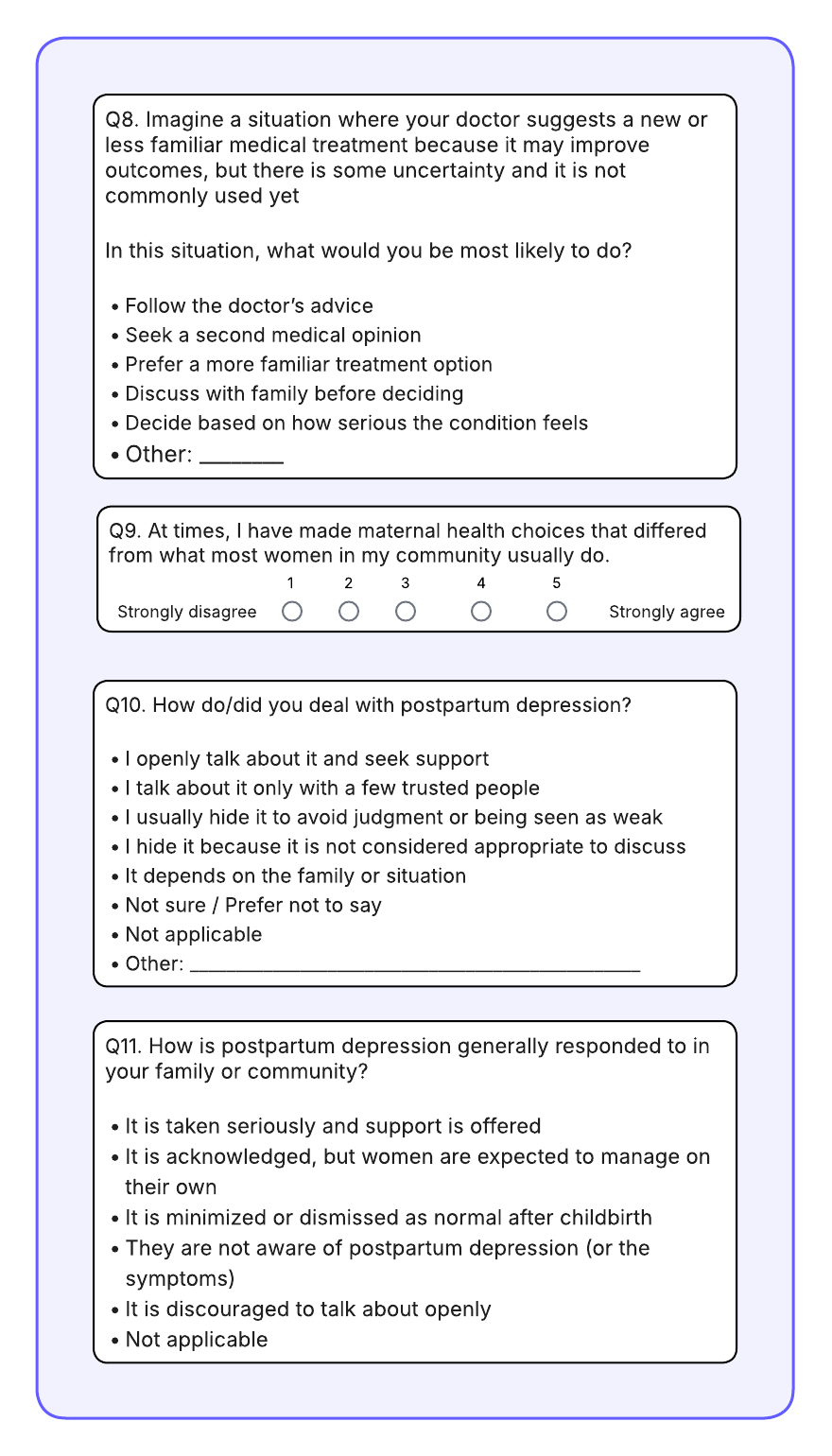}
\end{minipage}
\caption{The complete \framework{} survey instrument. Left: questions
1-7. Right: questions 8-11 (continued). Items combine Likert,
multiple-choice, and free-text response formats and map to the ten
\framework{} dimensions; the full item-to-dimension mapping is
provided in Appendix~\ref{app:scoring}.}
\label{fig:survey_instrument}
\end{figure*}

\begin{figure*}[t]
\centering
\begin{minipage}[t]{0.48\textwidth}
    \centering
    \includegraphics[width=\linewidth]{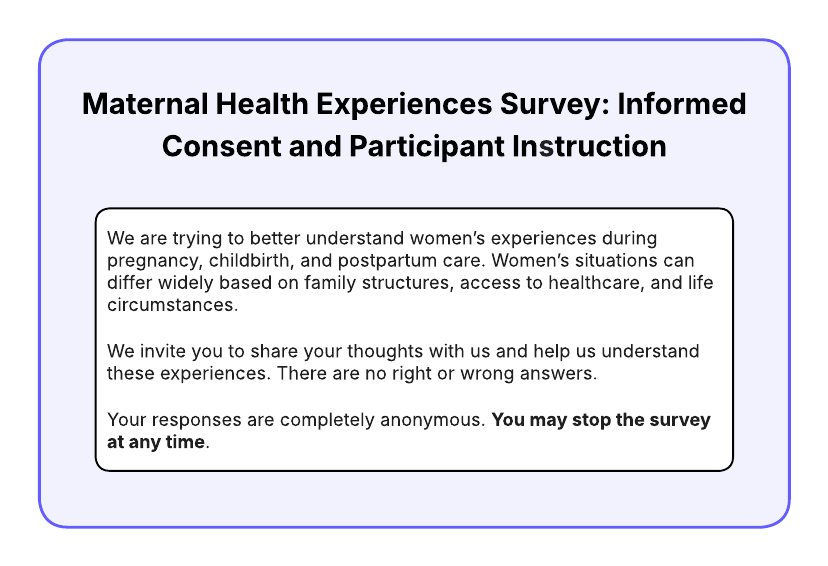}
\end{minipage}
\caption{Consent for \framework{} survey instrument.}
\label{fig:consent}
\end{figure*}

A snapshot of the survey administered to both human respondents and LLMs is
reproduced below Figure ~\ref{fig:survey_instrument} and Figure ~\ref{fig:consent}. Items operationalise the ten \framework{}
dimensions through a mix of Likert, multiple-choice, and free-text
formats.
\clearpage

\end{document}